%% file: main.tex
\def\isarxiv{1} %%% for icml submission version, we comment this line

\ifdefined\isarxiv
\documentclass[11pt]{article}

\usepackage[numbers]{natbib}

\else
\documentclass{article}
\usepackage{neurips_2022}
\fi

\usepackage{amsmath}
\usepackage{amsthm}
\usepackage{amssymb}
\usepackage{algorithm}
\usepackage{subfig}
\usepackage{algpseudocode}
\usepackage{graphicx}
\usepackage{grffile}
\usepackage{wrapfig,epsfig}
\usepackage{url}
\usepackage{xcolor}
\usepackage{epstopdf}

\usepackage{bbm}
\usepackage{dsfont}

\allowdisplaybreaks

\ifdefined\isarxiv

\usepackage{tikz}
\usepackage{hyperref}  %%% arxiv don't allow this.
\hypersetup{colorlinks=true,citecolor=blue,linkcolor=blue} %%% Zhao : maybe we should comment this in submission.
\usetikzlibrary{arrows}
\usepackage[margin=1in]{geometry}

\else

\usepackage{microtype}
\usepackage{hyperref}
\definecolor{mydarkblue}{rgb}{0,0.08,0.45}
\definecolor{mydarkgreen}{rgb}{0,0.8,0}
\hypersetup{colorlinks=true, citecolor=mydarkblue,linkcolor=mydarkblue}

\fi
\newtheorem{theorem}{Theorem}[section]
\newtheorem{lemma}[theorem]{Lemma}
\newtheorem{definition}[theorem]{Definition}

\newtheorem{fact}[theorem]{Fact}

\newcommand{\wt}{\widetilde}

\newcommand{\R}{\mathbb{R}}

\renewcommand{\d}{\mathrm{d}}

\DeclareMathOperator{\poly}{poly}

\DeclareMathOperator{\nnz}{nnz}

\DeclareMathOperator{\diag}{diag}

\DeclareMathOperator{\reg}{reg}

\makeatletter
\newcommand*{\RN}[1]{\expandafter\@slowromancap\romannumeral #1@}
\makeatother

\usepackage{lineno}

\begin{document}

\ifdefined\isarxiv

\date{}

\title{Solving Regularized Exp, Cosh and Sinh Regression Problems}
\author{
Zhihang Li\thanks{\texttt{lizhihangdll@gmail.com}. Huazhong Agricultural University.}
\and
Zhao Song\thanks{\texttt{zsong@adobe.com}. Adobe Research.}
\and
Tianyi Zhou\thanks{\texttt{t8zhou@ucsd.edu}. University of California San Diego.}
}

\else

\title{Intern Project} 
\maketitle 
\fi

\ifdefined\isarxiv
\begin{titlepage}
  \maketitle
  \begin{abstract}
\input{abstract}

  \end{abstract}
  \thispagestyle{empty}
\end{titlepage}

{\hypersetup{linkcolor=black}
\tableofcontents
}
\newpage

\else

\begin{abstract}
\input{abstract}
\end{abstract}

\fi

\input{intro} %%% Section 1. Introduction

\input{preli}

\newpage
\input{exp}

\newpage
\input{cosh}

\newpage
\input{sinh}

\newpage
\input{newton}

\ifdefined\isarxiv
%\section*{Acknowledgments}
\bibliographystyle{alpha}
\bibliography{ref}
\else
\bibliography{ref}
\bibliographystyle{alpha}

\fi

\newpage
\onecolumn
\appendix

%%%% Cut-line between first 10 pages and appendix

%%% some writing rules

%% Writing rule for creating tags.
%% Tags :
%% Theorem    \ref{thm:bla_bla}
%% Lemma      \ref{lem:bla_bla}
%% Claim      \ref{cla:bla_bla}
%% Corollary  \ref{cor:bla_bla}
%% Fact       \ref{fac:bla_bla}
%% Definition \ref{def:bla_bla}
%% Section    \ref{sec:bla_bla}
%% Subsection \ref{sub:bla_bla}
%% Equation   \ref{eq:bla_bla}

\end{document}

%% file: abstract.tex
In modern machine learning, attention computation is a fundamental task for training large language models such as Transformer, GPT-4 and ChatGPT.
In this work, we study the exponential regression problem which is inspired by the softmax/exp unit in the attention mechanism in large language models.  
The standard exponential regression is non-convex. We study the regularization version of the exponential regression problem which is a convex problem. We use the approximate newton method to solve in input sparsity time.

Formally, in this problem, one is given matrix $A \in \mathbb{R}^{n \times d}$, $b \in \mathbb{R}^n$, $w \in \mathbb{R}^n$ and any of functions $\exp, \cosh$ and $\sinh$ denoted as $f$. The goal is to find the optimal $x$ that minimize $ 0.5 \| f(Ax) - b \|_2^2 + 0.5 \| \diag(w) A x \|_2^2$. The straightforward method is to use the naive Newton's method. Let $\mathrm{nnz}(A)$ denote the number of non-zeros entries in matrix $A$. Let $\omega$ denote the exponent of matrix multiplication. Currently, $\omega \approx 2.373$. Let $\epsilon$ denote the accuracy error. In this paper, we make use of the input sparsity and purpose an algorithm that use $\log ( \|x_0  - x^*\|_2 / \epsilon)$ iterations and $\widetilde{O}(\mathrm{nnz}(A) + d^{\omega} )$ per iteration time to solve the problem.

%% file: intro.tex
\section{Introduction}

State-of-the-art language models like Transformer \cite{vsp+17}, BERT \cite{dclt18}, GPT-3 \cite{bmr+20}, PaLM \cite{cnd+22}, and OPT \cite{zrg+22} exhibit greater proficiency in natural language processing when compared to smaller models or traditional techniques. These models have the capacity to understand and generate intricate language, proving beneficial in various applications such as language translation, sentiment analysis, and question answering. LLMs can be customized for multiple purposes without necessitating their reconstruction from scratch. An instance of this is ChatGPT, an OpenAI-developed chat software that employs GPT-3's full potential. The latest iteration,  GPT-4 \cite{openai23},  has the potential to surpass GPT-3 in its impressive capabilities, including text generation, question answering, and language translation. This development could lead to significant implications in the field of NLP, with new applications potentially emerging in areas such as virtual assistants, chatbots, and automatic content creation. However, even though deep learning has a swift incline in popularity, we hold the belief that there exist discrepancies in our comprehension of the concept of attention and the reasoning behind its effectiveness.

The primary technical foundation behind LLMs is the attention matrix \cite{vsp+17,rns+18,dclt18,bmr+20,as23,zhdk23}. An attention matrix is a matrix that features rows and columns aligning with individual words or "tokens" and their relationships within a given text. Its purpose is to measure the critical nature of each token in a sequence in relation to the intended output. The attention matrix is learned during training. These parameters are optimized to maximize the model's accuracy in predicting the desired output. Through the attention mechanism, each input token is evaluated based on its importance or relevance to the desired output. This is achieved by weighing the token score, which is based on a similarity function comparing the current output state and input states.

More formally, the attention matrix can be expressed by considering two matrices,  $Q$ and $K$, containing query and key tokens, respectively. Both $Q$ and $K$ hold values in the $n \times d$ dimensional space. The attention matrix can be denoted by the square matrix $A$ which is of size $n \times n$. This matrix establishes a relationship between the input tokens in the sequence where every entry represents the attention weight or score between a particular input token (query token $Q$) and an output token (key token $K$). It is essential to mention that diagonal entries of this matrix display self-attention scores, signifying the importance of each token with respect to itself. A majority of the methods used for effective computation of attention matrices are divided into two primary categories based on their approach. One approach involves leveraging sparsity, as seen in Reformer \cite{kkll20}, while the other involves utilizing the low-rank attributes of the attention matrices, as observed in Linformer \cite{wlk+20} and Performer \cite{cld+20}. 

During training, our primary goal is to tackle the issue of multiple attention regression by utilizing the exponential function and its corresponding equation: $\min_{X} \|D^{-1} \exp(AX) - B \|_2$, where $D^{-1}$ denotes the normalization factor.
 However, upon further investigation, it has been brought to our attention that the single regression scenario has not been thoroughly studied. As a result, this study is centered on the situation of single regression. 

In our setting, the presence of $D$ is unnecessary due to the fact that $Ax$ is a column vector (but not a matrix). Thus, we have opted to address the optimization problem with regards to $\min_x \|\exp(Ax) - b\|_2$. It is important to note that our approach is not exclusive to the exponential function, and can also be extended to other hyperbolic functions such as $\cosh$ and $\sinh$.\footnote{$\cosh(x):=\frac{1}{2}(e^x + e^{-x})$ and $\sinh(x):=\frac{1}{2}(e^x - e^{-x})$}

\subsection{Our Results}
We state our result as follows.
\begin{theorem}[Main result, Informal version of Theorem~\ref{thm:main_formal}]\label{thm:main_informal}
Given matrix $A \in \R^{n \times d}$, $b \in \R^n$, and $w \in \R^n$. 

Let $f$ be any of functions $\exp, \cosh$ and $\sinh$. Let $g$ denote the gradient of function $f$.

Let $x^*$ denote the optimal solution of 
\begin{align*}
\min_{x \in \R^d} 0.5 \| f(Ax) - b \|_2^2 + 0.5 \| \diag(w) A x \|_2^2
\end{align*}
that $g(x^*) = {\bf 0}_d$  
and $\| x^* \|_2 \leq R$.

Let $\| A \| \leq R, \| b \|_2 \leq R$. Let $w_{i}^2 > 0.5b_{i}^2 + 1 + l/\sigma_{\min}(A)^2$ for all $i \in [n]$.

% \sqrt{n} 
Let $M = \exp(6 (R^2 +\log n))$.

Let $x_0$ denote an initial point such that $M \| x_0 - x^* \|_2 \leq 0.1 l$.

For any accuracy parameter $\epsilon \in (0,0.1)$ and failure probability $\delta \in (0,0.1)$.  There is a randomized algorithm (Algorithm~\ref{alg:main}) that runs $\log(\| x_0 - x^* \|_2/ \epsilon)$ iterations and spend 
\begin{align*}
O( (\nnz(A) + d^{\omega} ) \cdot \poly(\log(n/\delta)) 
\end{align*}
time per iteration, and finally outputs a vector $\wt{x} \in \R^d$ such that
\begin{align*}
\| \wt{x} - x^* \|_2 \leq \epsilon
\end{align*}
holds with probability at least $1-\delta$.
\end{theorem}

\begin{algorithm}[!ht]\caption{}\label{alg:main}
\begin{algorithmic}[1]
\Procedure{FastAlgorithm}{$A \in \R^{n \times d}, b \in \R^n, w \in \R^{n}, \epsilon \in (0,0.1), \delta \in (0,0.1)$} \Comment{Theorem~\ref{thm:main_informal}}
    \State Pick an initialization point $x_0$
    \State $T \gets \log( \| x_0 - x^* \|_2 / \epsilon )$
    \For{$t=0 \to T$}
        \State $D \gets \diag( (2 \exp( A x_{t} ) - b) \circ \exp( A x_{t} ) - w \circ w)$
        \State $\wt{D} \gets \textsc{SubSample}(D,A,\epsilon_1 = \Theta(1), \delta_1 = \delta/T)$ \Comment{Lemma~\ref{lem:subsample}}
        \State $g \gets A^\top \diag(\exp(Ax) \circ (\exp(Ax) - b) ) {\bf 1}_n$
        \State \Comment{$H = A^\top D A$ is the exat Hessian}
        \State \Comment{$x_{t+1} \gets x_t + H^{-1} g$ is the exact update}
        \State $\wt{H} \gets A^\top \wt{D} A$ \Comment{$\wt{H}$ is an approximation of $H$}
        \State $x_{t+1} \gets x_t + \wt{H}^{-1} g$ \Comment{This is an approximate update step}
    \EndFor
    \State $\wt{x}\gets x_{T+1}$
    \State \Return $\wt{x}$
\EndProcedure
\end{algorithmic}
\end{algorithm}

\subsection{Related Work}

\paragraph{Input sparsity time algorithms}

Input sparsity is a term used to describe datasets that have a majority of elements that are either zero or negligible. Utilizing algorithms optimized for sparse input data enables faster processing times than those utilized in dense data algorithms. This is because these algorithms can solely focus on non-zero elements, thereby minimizing computational and memory usage. 
Sparse algorithms’ time complexity depends only on the number of non-zero elements rather than the number of total elements. Input sparsity algorithms are highly applicable within fields such as solving John Ellipsoid \cite{syyz22}, discrepancy minimization \cite{dsw22},  low-rank approximation \cite{cw13,rsw16,swz17,swz19_soda}, subspace embeddings \cite{nn13}, column subset selection \cite{swz19_neurips1,swz19_neurips2} and  least squares regression \cite{dswy19}.

\paragraph{Algorithmic Regularization}
The standard exponential regression is non-convex. We study the regularization version of the exponential regression problem which is a convex problem.
In the context of deep learning, the non-convexity of the objective function necessitates the use of regularization techniques. Due to the possibility of the objective function generating multiple global minima that are widely scattered and vary significantly in their generalization capabilities, this becomes essential. Algorithmic regularization can be observed in various machine learning applications such as   binary classification  \cite{shn+18,ll19,cb20}, matrix factorization \cite{glss18,achl19}, convolutional neural networks \cite{glss18,jrg22}, generative adversarial networks \cite{azl21}, contrastive learning \cite{wl21} and mixture of experts \cite{cdw+22}. There are numerous factors that can induce algorithmic regularization. 
\cite {gdg+17, hhs17, kmn+16,
skyl18, lwm19} introduce how learning rate and batch size help regularize the training process. \cite{lwm19} explains that while a small initial learning rate may result in prompt training and better performance initially, a larger learning rate tends to yield improved generalization shortly after the annealing of the learning rate.  
The regularizer's role in the GAN model is expounded in \cite{azl21}. In addition, \cite{jl22} has conducted a theoretical analysis of the regularization generated by momentum. Furthermore, the employment of an adaptive step-size optimizer, such as Adam optimizer, can also function as a form of regularization during the training process \cite{kb14, nss15, d17, wrs+17,zclg21, jmgl22}. Batch normalization is analyzed in a theoretical capacity in \cite{ all18, hbgs19, is15}, which delves into its impact on regularization. A separate study conducted by \cite{shk+14, wkm20} explains how introducing dropout into the training process can prevent overfitting.
In \cite{dssw18}, they apply TensorSketch to the Kronecker product of multiple matrices efficiently without explicitly computing the tensor product and also apply the regularization to solve the Kronecker product regression.

\paragraph{Attention Theory}
 The topic of attention's expressivity has been a focus of early theoretical works. 
 In the case of self-attention blocks, \cite{vbc20} interpret self-attention as a system of self-interacting particles and theoretically explain the attention. 
 \cite{egkz21} explain it from inductive biases
and variable creation perspective. The role of attention in Transformers was studied by \cite{wcm21,dgv+18}. 
 In terms of optimization, \cite{zkv+20} examined the impact of adaptive approaches on attention models, while \cite{szks21} analyzed the dynamics of single-head attention to approximate Seq2Seq architecture's learning process.
For most LLMs,  it generally suffices to conduct attention computations in an approximate manner during the inference process, provided that there are adequate assurances of accuracy. Research conducted by various sources such as \cite{sparse_transformer19,kkll20,wlk+20,dkod20,kvpf20,cdw+21,cdl+22} has underscored this perspective. Due to that motivation,  \cite{zhdk23,as23} study the  computation of the attention matrix from the hardness perspective and purpose faster algorithms.

\paragraph{Newton Method and Hessian Computation}

Computing the Hessian or approximately computing the Hessian is a standard task in convex optimization. Many of the previous have work on this direction and use that improve several optimization problems such as linear programming \cite{cls19,lsz19,b20,blss20,sy21,jswz21,dly21,b21,gs22}, empirical risk minimization \cite{lsz19,qszz23}, cutting plane method \cite{jlsw20}, semi-definite programming \cite{jkl+20,hjs+22,gs22}, sum of squares \cite{jnw22}, training over-parameterized neural network \cite{zmg19, cgh+19,bpsw21,szz21,hswz22,z22}.

\paragraph{Roadmap.}
We organize the following paper as follows. In Section~\ref{sec:preli} we provide some tools for basic algebra and the analysis of a regularization term $L_{\reg}$. In Section~\ref{sec:exp} we provide detailed analysis of the loss function based on $\exp(x)$ (denoted as $L_{\exp}$) and the loss function with a regularization term $L_{\exp,\reg}$. In Section~\ref{sec:cosh} we provide detailed analysis of $L_{\cosh}$ and $L_{\cosh,\reg}$. In Section~\ref{sec:sinh} we provide detailed analysis of $L_{\sinh}$ and $L_{\sinh,\reg}$. In Section~\ref{sec:newton} we provide an approximate version of newton method which use for solving convex optimization problem which is more efficient under certain assumptions. 

%% file: preli.tex
\section{Preliminary}\label{sec:preli}
 
In this section, we provide preliminaries to be used in our paper. In Section~\ref{sec:preli:notations} we introduce notations we use. In Section~\ref{sec:preli:basic} we provide some facts about approximate computations and exact computations. In Section~\ref{sec:preli:gradient_hessian}, we provide some trivial facts regarding gradient and hessian. In Section~\ref{sec:preli:regularization}, we computed the gradient and hessian of a regularization term $L_{\reg}$.
\subsection{Notations}\label{sec:preli:notations}
We use $\R$ to denote real. We use $\R_{\geq 0}$ to denote non-negative real numbers.

For any vector $x \in \R^n$, we use $\exp(x) \in \R^n$ to denote a vector where $i$-th entry is $\exp(x_i)$. 

For a vector $x \in \R^n$, we use $\| x \|_2$ to denote its $\ell_2$ norm, i.e., $\| x \|_2 := ( \sum_{i=1}^n x_i^2 )^{1/2}$.

For any matrix $A \in \R^{n \times k}$, we denote the spectral norm of $A$ by $\| A \|$, i.e., 
\begin{align*}
\| A \| := \sup_{x\in\R^k} \| A x \|_2 / \| x \|_2.
\end{align*}

For a matrix $A$, we use $\sigma_{\max}(A)$ to denote the largest singular value of $A$. We use $\sigma_{\min}(A)$ to denote the smallest singular value of $A$.

For a vector $x \in \R^n$, we use $\| x \|_{\infty}$ to denote $\max_{i \in [n]} |x_i|$.

For each $x \in \R^n, y \in \R^n$, we use $x\circ y$ to denote a vector that has length $n$ and the $i$-th entry is $x_i y_i$ for all $i \in [n]$.

For a vector $x \in \R^n$ with each entry of $x_i$, we use $\diag(x)=A$ to generate a diagonal matrix $A \in \R^{n \times n}$ where each entry of $A$ on the diagonal is $A_{i,i}=x_i$ for all $i \in [n]$.

Given two column vectors $a, b \in \R^n$, we use $a \circ b$ to denote a column vector that $(a\circ b)_i$ is $a_ib_i$.

We use ${\bf 1}_n$ to denote a length-$n$ vector where all the entries are ones.

We say $A \succeq B$ if $x^\top  Ax \geq x^\top B x$ for all vector $x$.

We define $\cosh(x) = \frac{1}{2}( \exp(x) + \exp(-x))$ and $\sinh(x) = \frac{1}{2} ( \exp(x) - \exp(-x))$.

For any given matrix $A \in \R^{n \times d}$, we use $\nnz(A)$ to denote the number of non-zero entries of $A$, i.e., $\nnz(A) := | \{ (i,j) \in [n] \times [d] ~|~ A_{i,j} \neq 0 \} |$

For a diagonal matrix $D \in \R^{n \times n}$, we say $D$ is a $k$-sparse diagonal matrix, i.e., $k = |\{ i \in [n] ~|~ D_{i,i} \neq 0 \}|$.

For any function $f$, we use $\wt{O}(f)$ to denote $f \cdot \poly(\log f)$.

\subsection{Basic Algebras}\label{sec:preli:basic}

We state some facts which can give exact computation.
\begin{fact}[$\exp$, $\cosh$, $\sinh$ exact and approximate for general range]\label{fac:e_cosh_sinh_exact}
We have 
\begin{itemize}
    \item $\frac{\d \exp(x) }{ \d x} = \exp(x)$
    \item $\frac{\d \cosh(x)}{ \d x } = \sinh(x)$
    \item $\frac{\d \sinh(x)}{ \d x} = \cosh(x)$
    \item $\cosh^2(x) - \sinh^2(x) = 1$
    
    \item $|\exp(x)| \leq \exp(|x|)$
    \item $|\cosh(x)| = \cosh(|x|) \leq \exp(|x|)$  
    \item $|\sinh(x)| =  \sinh(|x|)$ 
    \item $\exp(x) = \sum_{i=0}^{\infty} \frac{1}{i!} x^i$
    \item $\cosh(x) = \sum_{i=0}^{\infty} \frac{1}{(2i)!} x^{2i} $
    \item $\sinh(x) = \sum_{i=0}^{\infty} \frac{1}{(2i+1)!} x^{2i+1} $
\end{itemize}
\end{fact}

We state some facts which can give a reasonable approximate computation.
\begin{fact}[$\exp$, $\cosh$, $\sinh$ approximate computation in small range]\label{fac:e_cosh_sinh_approx}
We have
\begin{itemize}
    \item Part 1. For any $x$ satisfy that $ |x| \leq 0.1$, we have $|\exp(x) - 1| \leq 2|x|$
    \item Part 2. For any $x$ satisfy that $|x| \leq 0.1$, we have $|\cosh(x) - 1| \leq x^2$
    \item Part 3. For any $x$ satisfy that $|x| \leq 0.1$, we have $|\sinh(x) | \leq 2|x|$
    \item Part 4. For any $x,y$ satisfy that $| x - y | \leq 0.1$, we have $| \exp(x) - \exp(y) | \leq \exp(x) \cdot 2 |x-y|$
    \item Part 5. For any $x,y$ satisfy that $| x - y | \leq 0.1$, we have $| \cosh(x) - \cosh(y) | \leq \cosh(x) \cdot 2|x - y|$
    \item Part 6. For any $x,y$ satisfy that $| x - y | \leq 0.1$, we have $| \sinh(x) - \sinh(y) | \leq \cosh(x) \cdot 2|x - y|$
\end{itemize}
\end{fact}
\begin{proof}
Most of the proofs are standard, we only provide proofs for some of them.

{\bf Proof of Part 5.}

We have
\begin{align*}
|\cosh(x) - \cosh(y)|
= & ~ | 0.5 \exp(x) + 0.5 \exp(-x) - 0.5 \exp(y) - 0.5 \exp(-y)| \\
\leq & ~ | 0.5 \exp(x) - 0.5 \exp(y) | + | 0.5 \exp(-x) - 0.5 \exp(-y)| \\
\leq & ~ 0.5 \exp(x) |1-\exp(y-x)| + 0.5 \exp(-x) |1-\exp(-y+x)| \\
\leq & ~ 0.5 \exp(x) \cdot 2 |y-x| + 0.5 \exp(-x)  \cdot 2 |y-x| \\
= & ~ \cosh(x) \cdot 2|x-y|
\end{align*}
where the first step follows from the definition of $\cosh$, the second step follows from triangle inequality, the third step follows from simple algebra,
 the fourth step follows from the fact that $|\exp(x) - 1| \leq 2x$ for all $x \in [0,0.1]$ and the last step follows from the definition of $\cosh$.

 {\bf Proof of Part 6.}

We have
 \begin{align*}
| \sinh(x) - \sinh(y) | = & ~ | 0.5 \exp (x) - 0.5 \exp(-x) - 0.5 \exp(y) + 0.5 \exp( -y) |\\
\leq & ~ | 0.5 \exp(x) - 0.5 \exp(y) | + |0.5 \exp(-y) - 0.5\exp(-x)| \\
= & ~ 0.5 \exp(x) |1 - \exp(y-x)| + 0.5 \exp(-x) |\exp(x-y) - 1| \\
\leq & ~ 0.5 \exp(x) \cdot 2 |y-x| + 0.5 \exp(-x)  \cdot 2 |y-x| \\
= & ~ \cosh(x) \cdot 2|x-y|
\end{align*}
where the first step follows from the definition of $\sinh$, the second step follows from triangle inequality, the third step follows from simple algebra,
 the fourth step follows from the fact that $|\exp(x) - 1| \leq 2x$ for all $x \in [0,0.1]$ and the last step follows from the definition of $\cosh$.
\end{proof}

\begin{fact}\label{fac:circ_diag}
For any vectors $a \in \R^n, b \in \R^n$ and $c \in \R^n$, we have 
\begin{itemize}
    \item $a \circ b = b \circ a = \diag(a) \cdot b = \diag(b) \cdot a = \diag(a) \cdot \diag(b) \cdot {\bf 1}_n $
    \item $a^\top (b \circ c)= a^\top \diag(b) c$
    \begin{itemize}
        \item $a^\top (b \circ c) = b^\top (a \circ c) = c^\top (a \circ b)$
        \item $a^\top \diag(b) c = b^\top \diag(a) c = a^\top \diag(c) b$
    \end{itemize}
    \item $\diag(a \circ b) = \diag(a) \diag(b)$
    \item $\diag(a) + \diag(b) = \diag(a + b) $
\end{itemize}
\end{fact}

\begin{fact}\label{fac:vector_norm}
For vectors $a,b \in \R^n$, we have
\begin{itemize}
    \item Part 1. $\| a \circ b \|_2 \leq \| a \|_{\infty} \cdot \| b \|_2$
    \item Part 2. $\| a \|_{\infty} \leq \| a \|_2 \leq \sqrt{n} \cdot \| a \|_{\infty}$
    \item Part 3. $\| \exp(a) \|_{\infty} \leq \exp(\| a \|_2)$
    \item Part 4. $\| \cosh(a) \|_{\infty} \leq \cosh( \| a \|_2 ) \leq \exp( \| a \|_2 )$
    \item Part 5. $\| \sinh(a) \|_{\infty} \leq \sinh(\| a \|_2) \leq \cosh(\| a \|_2) \leq \exp(\| a \|_2)$
    \item Part 6. $\cosh(a)\circ \cosh(a) - \sinh(a) \circ \sinh(a) = {\bf 1}_n$
    \item Part 7. For any $\| a - b \|_{\infty} \leq 0.01$, we have $\| \exp(a) - \exp(b) \|_2 \leq \| \exp(a) \|_2 \cdot 2 \| a - b \|_{\infty}$
    \item Part 8. For any $\| a - b \|_{\infty} \leq 0.01$, we have $\| \cosh(a) - \cosh(b) \|_2 \leq \| \cosh(a) \|_2 \cdot 2 \| a - b \|_{\infty}$
    \item Part 9. For any $\| a - b \|_{\infty} \leq 0.01$, we have $\| \sinh(a) - \sinh(b) \|_2 \leq \| \cosh(a) \|_2 \cdot 2 \| a - b \|_{\infty}$
\end{itemize}
\end{fact}
\begin{proof}
Most of the proofs are standard, we only provide proofs for some of them.

{\bf Proof of Part 8.}

We have
\begin{align*}
    | \cosh (a_i) - \cosh (b_i) | \leq & ~ | \cosh (a_i) | \cdot  2 | a_i -b_i | \\
    \leq & ~  | \cosh (a_i) | \cdot 2 \| a - b\| _\infty ,
\end{align*}
where the first step follows from Part 5 in Fact \ref{fac:e_cosh_sinh_approx} and the last step follows from the definition of norm.
By summing up the square of both sides, we have
\begin{align*}
    \| \cosh(a) - \cosh(b) \|_2 \leq \|\cosh(a)\|_2 \cdot 2\|a-b\|_\infty
\end{align*}

{\bf Proof of Part 9.}

We have
\begin{align*}
    | \sinh (a_i) - \sinh (b_i) | \leq & ~ | \cosh (a_i) | \cdot  2 | a_i - b_i| \\
    \leq & ~  | \cosh (a_i) | \cdot 2 \| a - b\| _\infty ,
\end{align*}
where the first step follows from Part 6 in Fact \ref{fac:e_cosh_sinh_approx} and the last step follows from the definition of norm.
By summing up the square of both sides, we have
\begin{align*}
    \| \sinh(a) - \sinh(b) \|_2 \leq \|\cosh(a)\|_2 \cdot 2\|a-b\|_\infty
\end{align*}
\end{proof}

\begin{fact}\label{fac:matrix_norm}
For matrices $A,B$, we have 
\begin{itemize}
    \item $\| A^\top \| = \| A \|$
    \item $\| A \| \geq \| B \| - \| A - B \|$
    \item $\| A + B \| \leq \| A \| + \| B \|$
    \item $\| A \cdot B \| \leq \| A \| \cdot \| B \|$ 
    \item If $A \preceq \alpha \cdot B$, then $\| A \| \leq \alpha \cdot \| B \|$
\end{itemize}
\end{fact}

\begin{fact}\label{fac:psd}
Let $\epsilon \in (0,0.1)$. 
If 
\begin{align*}
(1-\epsilon) A \preceq B \preceq (1+\epsilon) A
\end{align*}
Then
 
\begin{itemize}
    \item $-\epsilon A \preceq B- A \preceq  \epsilon A$
    \item $-\epsilon \preceq A^{-1/2} (B-A) A^{-1/2} \preceq \epsilon $
    \item $\| A^{-1/2} (B-A) A^{-1/2} \| \leq \epsilon$
    \item $\| A^{-1}  ( B - A ) \|\leq \epsilon$
    \item $(1+\epsilon)^{-1} B \preceq A \preceq (1-\epsilon)^{-1} A $
    \item $(1-2\epsilon) B \preceq A \preceq (1+2\epsilon) B $
    \item $\| B^{-1/2} (B-A) B^{-1/2} \| \leq 2\epsilon$
    \item $\| B^{-1} (B-A) \| \leq 2\epsilon$
\end{itemize}
\end{fact}

\subsection{Standard Gradient and Hessian Computation}\label{sec:preli:gradient_hessian}

\begin{lemma}[Standard Gradient and Hessian Computation]\label{lem:Ax_gradient_hessian}
We have
\begin{itemize}
\item Part 1. $ \frac{\d Ax}{\d t} = A \frac{\d x}{ \d t}$.
\item Part 2. $  \frac{\d Ax}{\d x_i} = A_{*,i} \in \R^n $
\item Part 3.  $ \frac{\d^2 Ax}{\d x_i^2}  =  0 $
\end{itemize}
\end{lemma}
\begin{proof}

{\bf Proof of Part 1.}

It trivially follows from chain rule.

{\bf Proof of Part 2.}

The equation takes derivative of the vector $x$ by each entry $x_i$ of itself and trivially gets the result of $A_{*,i}$.

{\bf Proof of Part 3.}
\begin{align*}
        \frac{\d^2 Ax}{\d x_i^2}
        = & ~ \frac{\d}{\d x_i}\bigg(\frac{\d Ax}{\d x_i} \bigg) \\
        = & ~ \frac{\d A_{*,i}}{\d x_i} \\
        = & ~ 0
    \end{align*}
    where the first step is an expansion of the Hessian, the second step follows from the differential chain rule, and the last step is due to the constant entries of the matrix $A_{*,i}$.

\end{proof}

\subsection{Regularization term}\label{sec:preli:regularization}

\begin{lemma}\label{lem:regularization}
Let $w \in \R^n$ denote a weight vector
Let $L_{\reg}:=0.5 \| W A x \|_2^2$. Let $W = \diag(w) \in \R^{n \times n}$ denote a diagonal matrix.

 Then we have
\begin{itemize}
\item Part 1. Gradient
\begin{align*}
\frac{\d L_{\reg}}{\d x} = A^\top W^2 A x
\end{align*}
\item Part 2. Hessian
\begin{align*}
\frac{ \d^2 L_{\reg} }{\d x^2} = A^\top W^2 A
\end{align*}
\end{itemize}
\end{lemma}

\begin{proof}
{\bf Proof of Part 1.}
\begin{align*}
    \frac{\d L_{\reg}}{\d x}
    = & ~ ( \frac{\d}{\d x} (WAx ) )^\top \cdot  (WAx) \\
    = & ~ A^\top W^\top \cdot WAx  \\
    = & ~ A^\top W^2 Ax
\end{align*}
where the first step follows from chain rule.

{\bf Proof of Part 2.}
\begin{align*}
    \frac{ \d^2 L_{\reg} }{\d x^2}
    = & ~ \frac{\d}{\d x}\Big(\frac{ \d L_{\reg} }{\d x} \Big) \\
    = & ~ \frac{\d}{\d x}\Big(A^\top W^2 Ax \Big) \\
    = & ~ A^\top W^2 A
\end{align*}
where the first step follows from the expansion of Hessian, the second step follows from by applying the arguments in Part 1, the third step follows from simple algebra.
\end{proof}

\begin{lemma}\label{lem:ADA_pd}
Let $A \in \R^{n \times d}$.

Let $D \in \R^{n \times n}$ denote a diagonal matrix where all diagonal entries are positive. Then we have
\begin{align*}
A^\top D A  \succeq \sigma_{\min}(A)^2 \cdot \min_{i \in [n]} D_{i,i}  \cdot I_d
\end{align*}
\end{lemma}
\begin{proof}

For any $x \in \R^d$, we have
\begin{align*}
  x^\top A^\top D A x 
= & ~ \| \sqrt{D} A x \|_2^2 \\
= & ~ \sum_{i=1}^n D_{i,i} (Ax)_i^2 \\
\geq & ~ ( \min_{i \in [n]} D_{i,i} ) \cdot \sum_{i=1}^n (Ax)_i^2 \\
\geq & ~ \min_{i \in [n]} D_{i,i} \cdot \| A x \|_2^2 \\
\geq & ~ \min_{i \in [n]} D_{i,i} \cdot \sigma_{\min}(A)^2 \cdot \| x \|_2^2 \\
= & ~ x^\top \cdot ( \min_{i \in [n]} D_{i,i} \cdot \sigma_{\min}(A)^2 \cdot  I_d ) x
\end{align*} 
where the first step follows from simple algebra,
the second step follows from the definition of $\|x\|_2$,
the third step follows from simple algebra,
the fourth step follows from the definition of $\|x\|_2$,
the fifth step follows from 
definition of $\sigma_{\min}(A)$
, the last step follows from 
simple algebra
.
\end{proof}

%% file: exp.tex
\section{Exponential Regression}\label{sec:exp}
In this section, we provide detailed analysis of $L_{\exp}$. In Section~\ref{sec:exp:definition} we define the loss function $L_{\exp}$ based on $\exp(x)$. In Section~\ref{sec:exp:gradient} we compute the gradient of $L_{\exp}$ by detail. In Section~\ref{sec:exp:hessian} we compute the hessian of $L_{\exp}$ by detail. In Section~\ref{sec:exp:gradient_hessian}, we summarize the result of Section~\ref{sec:exp:gradient} and Section~\ref{sec:exp:hessian} and aquire the gradient $\nabla L_{\exp}$ and hessian $\nabla^2 L_{\exp}$ for $L_{\exp}$. In Section~\ref{sec:exp:loss_reg} we define $L_{\exp,\reg}$ by adding the regularization term $L_{\reg}$ in Section~\ref{sec:preli:regularization} to $L_{\exp}$ and compute the gradient $\nabla L_{\exp,\reg}$ and hessian $\nabla^2 L_{\exp,\reg}$ of $L_{\exp,\reg}$. In Section~\ref{sec:exp:convex} we proved that $\nabla^2 L_{\exp,\reg} \succ 0$ and thus showed that $L_{\exp,\reg}$ is convex. In Section~\ref{sec:exp:lipschitz} we provide the upper bound for $\|\nabla^2 L_{\exp,\reg}(x)-\nabla^2 L_{\exp,\reg}(y)\|$ and thus proved $\nabla^2 L_{\exp,\reg}$ is lipschitz.
\subsection{Definitions}\label{sec:exp:definition}

\begin{definition}[Loss function for Exp Regression]\label{def:L_exp}
Given $A \in \R^{n \times d}$ and $b \in \R^n$. 
For a vector $x \in \R^d$, we define loss function $L_{\exp}(x)$ as follows
\begin{align*}
L_{\exp}(x):= 0.5 \cdot \| \exp(Ax) - b \|_2^2
\end{align*}
\end{definition}

\subsection{Gradient}\label{sec:exp:gradient}

\begin{lemma}[Gradient for Exp]\label{lem:gradient_exp}
We have
\begin{itemize}
    
    \item Part 1.
    \begin{align*}
        \frac{ \d ( \exp(Ax) - b ) }{ \d t } = \exp(Ax) \circ \frac{A \d x }{ \d t}
    \end{align*}
    \item Part 2. 
    \begin{align*}
        \frac{\d L_{\exp} }{ \d t} = (\exp(Ax) - b )^\top \cdot ( \exp(Ax) \circ \frac{A \d x}{ \d t } )
    \end{align*}
\end{itemize}
Further, we have for each $i \in [d]$
\begin{itemize}
    
    \item Part 3.
    \begin{align*}
        \frac{ \d ( \exp(Ax) - b ) }{ \d x_i } = \exp(Ax) \circ A_{*,i}
    \end{align*}
    \item Part 4. 
    \begin{align*}
        \frac{\d L_{\exp} }{ \d x_i}  
        = & ~ (\exp(Ax) - b )^\top \cdot ( \exp(Ax) \circ A_{*,i} )
    \end{align*}
    \item Part 5.
    \begin{align*}
        \frac{\d L_{\exp} }{ \d x} =  A^\top \diag( \exp(Ax) ) (\exp(Ax) - b)
    \end{align*}
\end{itemize}
\end{lemma}

\begin{proof}

{\bf Proof of Part 1.}

For each $i \in [n]$, we have  
\begin{align*}
 \frac{ \d ( \exp(Ax) - b )_i }{ \d t } 
 = & ~ \exp(Ax)_i \cdot \frac{\d (Ax)_i}{\d t} \\
 = & ~ \exp(Ax)_i \cdot \frac{ (A \d x)_i}{\d t} 
\end{align*}
where the first and second step follow from the differential chain rule.

%Thus, we complete the proof.

{\bf Proof of Part 2.}

We have
 
\begin{align*}
    \frac{\d L_{\exp} }{ \d t} 
    = & ~ (\exp(Ax) - b )^\top \cdot \frac{ \d ( \exp(Ax) - b ) }{\d t} \\
    = & ~ (\exp(Ax) - b )^\top \cdot ( \exp(Ax) \circ \frac{A \d x}{ \d t } )
\end{align*}
where the first step follows from the differential chain rule, and the second step follows from $\frac{ \d ( \exp(Ax) - b ) }{ \d t } = \exp(Ax) \circ \frac{A \d x }{ \d t}$ in {\bf Part 1}.

{\bf Proof of Part 3.}

We have
\begin{align*}
    \frac{\d (\exp(Ax)-b)}{\d x_i}
    = & ~ \frac{\d (\exp(Ax))}{\d x_i} - \frac{\d b}{\d x_i}\\
    = & ~ \exp(Ax) \circ \frac{\d Ax}{\d x_i} - 0\\
    = & ~ \exp(Ax) \circ A_{*,i}
\end{align*}
where the first step follows from the property of the gradient, the second step follows from simple algebra, and the last step directly follows from Lemma~\ref{lem:Ax_gradient_hessian}.

{\bf Proof of Part 4.}

By substitute $x_i$ into $t$ of Part 3, we get
\begin{align*}
    \frac{\d L}{\d x_i}
    = & ~ (\exp(Ax) - b )^\top \cdot ( \exp(Ax) \circ \frac{A \d x}{ \d x_i } )\\
    = & ~ (\exp(Ax) - b )^\top \cdot ( \exp(Ax) \circ A_{*,i} ) \\
    = & ~ A_{*,i}^\top \diag( \exp(Ax) ) (\exp(Ax) - b )
\end{align*}
where the first step follows from the result of {\bf Part 2} and $\frac{\d y^2}{\d x} = 2y^\top \frac{\d y}{\d x}$,
the second step follows from the result of Lemma~\ref{lem:Ax_gradient_hessian}, 
the last step follows from Fact~\ref{fac:circ_diag}.

{\bf Proof of Part 5.}

We have
\begin{align*}
    \frac{\d L}{ \d x}
    = & ~ A^\top \diag( \exp(Ax) ) (\exp(Ax) - b)
\end{align*}
where this step follows from the result of {\bf Part 4} directly.
\end{proof}

\subsection{Hessian}\label{sec:exp:hessian}

\begin{lemma}\label{lem:hessian_exp}
\begin{itemize}

    \item Part 1.
    \begin{align*}
        \frac{ \d^2 ( \exp(Ax) - b ) }{ \d x_i^2 }
        = & ~ A_{*,i} \circ \exp(Ax) \circ A_{*,i}
    \end{align*}
    \item Part 2.
    \begin{align*}
        \frac{ \d^2 ( \exp(Ax) - b ) }{ \d x_i \d x_j }
        = & ~ A_{*,j} \circ \exp(Ax) \circ A_{*,i}
    \end{align*}
    \item Part 3. 
    \begin{align*}
        \frac{\d^2 L_{\exp} }{ \d x_i^2}
        = & ~ A_{*,i}^\top \diag( 2 \exp(Ax) - b ) \diag(\exp(Ax)) A_{*,i}
    \end{align*}
    \item Part 4. 
     \begin{align*}
        \frac{\d^2 L_{\exp} }{ \d x_i \d x_j} = A_{*,i}^\top \diag( 2 \exp(Ax) - b ) \diag(\exp(Ax)) A_{*,j}
    \end{align*}
\end{itemize}
\end{lemma}
\begin{proof}

{\bf Proof of Part 1.}
\begin{align*}
    \frac{ \d^2 ( \exp(Ax) - b ) }{ \d x_i^2 }
        = & ~ \frac{\d}{\d x_i}\bigg(\frac{\d (\exp(Ax) - b)}{\d x_i} \bigg) \\
        = & ~ \frac{\d (\exp(Ax) \circ A_{*,i})}{\d x_i} \\
        = & ~ A_{*,i} \circ \frac{\d \exp(Ax)}{\d x_i} \\
        = & ~ A_{*,i} \circ \exp(Ax) \circ A_{*,i}
\end{align*}
where the first step is an expansion of the Hessian, the second step follows from the differential chain rule, the third step extracts the matrix $A_{*,i}$ with constant entries out of the derivative, and the last step also follows from the chain rule.

{\bf Proof of Part 2.}
\begin{align*}
    \frac{ \d^2 ( \exp(Ax) - b ) }{ \d x_i \d x_j }
        = & ~ \frac{\d}{\d x_i}\bigg(\frac{\d}{\d x_j}\bigg(\exp(Ax) - b\bigg) \bigg) \\
        = & ~ \frac{\d}{\d x_i}\bigg(\exp(Ax) \circ A_{*,j} \bigg) \\
        = & ~ A_{*,j} \circ \exp(Ax) \circ A_{*,i}
\end{align*}
where the first step is an expansion of the Hessian, 
the second step follows from {\bf Part 3} of Lemma~\ref{lem:gradient_exp},
the third step follows from {\bf Part 3} of Lemma~\ref{lem:gradient_exp}.

{\bf Proof of Part 3.}
\begin{align*}
    \frac{\d^2 L }{ \d x_i^2}
        = & ~ \frac{\d}{\d x_i}\bigg(\frac{\d L}{\d x_i} \bigg) \\
        = & ~ \frac{\d}{\d x_i}\bigg((\exp(Ax) - b )^\top \cdot ( \exp(Ax) \circ A_{*,i} ) \bigg) \\
        = & ~ (\exp(Ax) \circ A_{*,i})^\top \cdot ( \exp(Ax) \circ A_{*,i} )+(\exp(Ax) - b )^\top \cdot (A_{*,i} \circ \exp(Ax) \circ A_{*,i})\\ 
        = & ~ A_{*,i}^\top \diag( 2 \exp(Ax) - b ) \diag(\exp(Ax)) A_{*,i}
\end{align*}
where the first step is an expansion of the Hessian,
the second step follows from {\bf Part 4} of Lemma~\ref{lem:gradient_exp},
the third step follows from differential chain rule and {\bf Part 1} of Lemma~\ref{lem:gradient_exp},
the last step follows from Fact~\ref{fac:circ_diag}.

{\bf Proof of Part 4.}
\begin{align*}
    \frac{\d^2 L }{ \d x_i \d x_j}
        = & ~ \frac{\d}{\d x_i}\bigg(\frac{\d L}{\d x_j} \bigg) \\
        = & ~ \frac{\d}{\d x_i}\bigg((\exp(Ax) - b )^\top \cdot ( \exp(Ax) \circ A_{*,j} ) \bigg) \\
        = & ~ (\exp(Ax) \circ A_{*,i})^\top \cdot ( \exp(Ax) \circ A_{*,j} )+(\exp(Ax) - b )^\top \cdot (A_{*,j} \circ \exp(Ax) \circ A_{*,i}) \\ 
        = & ~ A_{*,i}^\top \diag( 2 \exp(Ax) - b ) \diag(\exp(Ax)) A_{*,j}
\end{align*}
where the first step is an expansion of the Hessian, 
the second step follows from {\bf Part 4} of Lemma~\ref{lem:gradient_exp},
the third step follows from differential chain rule and {\bf Part 1} of Lemma~\ref{lem:gradient_exp}, 
the last step follows from Fact~\ref{fac:circ_diag}.  

\end{proof}

\subsection{Gradient and Hessian of the Loss function for Exp Function}\label{sec:exp:gradient_hessian}

\begin{lemma}\label{lem:gradient_hessian_exp}
    Let $L_{\exp}: \R^d \to \R_{\geq 0}$ be defined in Definition~\ref{def:L_exp}. Then for any $i, j \in [d]$, we have  
    \begin{itemize}
        \item Part 1. Gradient
    \begin{align*}
        \nabla L_{\exp} = A^\top \diag(\exp(Ax)) \diag(\exp(Ax) - b) {\bf 1}_n
    \end{align*}
        \item Part 2. Hessian
        \begin{align*}
            \nabla^2 L_{\exp} = A^\top \diag(2 \exp(Ax) - b) \diag( \exp(Ax) ) A
        \end{align*}
    \end{itemize}
\end{lemma}
 
\begin{proof}

{\bf Part 1.}
We run Lemma~\ref{lem:gradient_exp} and Fact~\ref{fac:circ_diag} directly.

{\bf Part 2.}
It follows from Part 3 and 4 of Lemma~\ref{lem:hessian_exp}. 
\end{proof}

\subsection{Loss Function with a Regularization Term}\label{sec:exp:loss_reg}

\begin{definition}\label{def:L_exp_and_regularized}
Given matrix $A \in \R^{n \times d}$ and $b \in \R^n$, $w \in \R^n$. For a vector $x \in \R^d$, we define loss function $L_{\exp,\reg}(x)$ as follows
\begin{align*}
L_{\exp,\reg}(x): = 0.5 \cdot \| \exp(Ax) - b \|_2^2+ 0.5 \cdot \| W A x \|_2^2
\end{align*}
where $W = \diag(w)$.
\end{definition}

\begin{lemma}
Let $L_{\exp,\reg}$ be defined as Definition~\ref{def:L_exp_and_regularized}, then we have
\begin{itemize}
    \item Part 1. Gradient
    \begin{align*}
        \frac{\d L_{\exp,\reg}}{\d x} = A^\top \diag(\exp(Ax)) \diag(\exp(Ax) - b) {\bf 1}_n + A^\top W^2 A x
    \end{align*}
    \item Part 2. Hessian
    \begin{align*} 
        \frac{\d^2 L_{\exp,\reg}}{\d x^2} = A^\top \diag(2 \exp(Ax) - b) \diag(\exp(Ax)) A +  A^\top W^2 A
    \end{align*}
    
\end{itemize}
\end{lemma}
\begin{proof}
{\bf Proof of Part 1.}
We run Lemma~\ref{lem:gradient_hessian_exp} and Lemma~\ref{lem:regularization} directly.

{\bf Proof of Part 2.}
We run Lemma~\ref{lem:gradient_hessian_exp} and Lemma~\ref{lem:regularization} directly.

\end{proof}

\subsection{Hessian is Positive Definite}\label{sec:exp:convex}

\begin{lemma}[Hessian is positive definite]\label{lem:hessian_is_pd_exp}
Let $l > 0 $ denote a parameter.

If the following condition hold
\begin{itemize}
    \item Let $H(x) = \frac{\d^2 L_{\exp,\reg}}{\d x^2}$
    \item $w_{i}^2 > 0.5 b_{i}^2 + l/\sigma_{\min}(A)^2 $ for all $i \in [n]$
\end{itemize}
Then, we have  
    \begin{align*}
         H(x) \succeq l \cdot I_d
    \end{align*}
   
\end{lemma}
\begin{proof}
We define $D$
\begin{align*}
D = \diag(2 \exp(Ax) - b ) \diag(\exp(Ax)) + W^2
\end{align*}

Then we can rewrite Hessian as 
\begin{align*}
\frac{\d^2 L}{\d x^2} = A^\top D A.
\end{align*} 

We define
\begin{align*}
z = \exp(Ax).
\end{align*}
 
Then we have
\begin{align*}
D_{i,i} 
= & ~ 2( \exp( (Ax)_i ) - b_i ) \exp( (Ax)_i ) + w_{i,i}^2 \\
= & ~ 2 (z_i - b_i) z_i + w_{i,i}^2 \\
> & ~ 2 (z_i - b_i) z_i + 0.5 b_{i}^2 + l/\sigma_{\min}(A)^2 \\
= & ~ 0.5 ( 2z_i - b_i )^2 + l/\sigma_{\min}(A)^2 \\
\geq & ~ l/\sigma_{\min}(A)^2
\end{align*}
where the first step follows from simple algebra, the second step follows from replacing $\exp(Ax)$ with $z = \exp(Ax)$, the third step follows from $w_{i}^2 > 0.5b_{i}^2 + l/\sigma_{\min}(A)^2$, the fourth step follows from simple algebra, the fifth step follows from $x^2\geq0, \forall x$.

Since we know $D_{i,i} > l/\sigma_{\min}(A)^2$ for all $i \in [n]$ and Lemma~\ref{lem:ADA_pd}, we have 
\begin{align*}
A^\top D A \succeq (\min_{i \in [n]} D_{i,i}) \cdot \sigma_{\min}(A)^2 I_d \succeq l \cdot I_d
\end{align*}
Thus, Hessian is positive definite forever and thus the function is convex.

\end{proof}

\subsection{Hessian is Lipschitz}\label{sec:exp:lipschitz}
\begin{lemma}[Hessian is Lipschitz]\label{lem:hessian_is_lipschitz_exp}
If the following condition holds
 
\begin{itemize}
    \item Let $H(x) = \frac{\d^2 L_{\exp,\reg}}{\d x^2}$
    \item Let $R > 2$
    \item $\|x \|_2 \leq R, \| y \|_2 \leq R$
    \item $\| A (x-y) \|_{\infty} < 0.01$
    \item $\| A \| \leq R$
    \item $\| b \|_2 \leq R$
\end{itemize}
Then we have
    \begin{align*}
        \| H(x) - H(y) \| \leq \sqrt{n} \cdot \exp(6 R^2) \cdot \| x - y \|_2
    \end{align*}
\end{lemma}
\begin{proof}

We have
\begin{align}\label{eq:rewrite_H_diff_exp}
& ~ \| H(x) - H(y) \| \notag \\
= & ~ \| A^\top \diag(2 \exp(Ax) - b) \diag(\exp(Ax)) A -  A^\top \diag(2 \exp(Ay) - b) \diag(\exp(Ay)) A \| \notag \\
\leq & ~ \| A \|^2 \cdot \|  (2 \exp(Ax) - b) \circ \exp(Ax) - (2 \exp(Ay) - b) \circ \exp(Ay) \|_2  \notag \\
= & ~ \| A \|^2 \cdot \| 2 (\exp(Ax) + \exp(Ay) )\circ ( \exp(Ax) - \exp(Ay) ) - b \circ ( \exp(Ax) - \exp(A y) ) \|_2 \notag \\
= & ~ \| A \|^2 \cdot \| ( 2 \exp(Ax) + 2 \exp(Ay) - b ) \circ ( \exp(Ax) - \exp(Ay) ) \|_2  \notag \\
\leq & ~ \| A \|^2 \cdot \| ( 2 \exp(Ax) + 2 \exp(Ay) - b ) \|_{\infty} \cdot \| \exp(Ax) - \exp(Ay) \|_2
\end{align}
where the first step follows from $H(x) = \nabla^2L$ and simple algebra, the second step follows from Fact~\ref{fac:matrix_norm}, the third step follows from simple algebra, the fourth step follows from simple algebra, the last step follows from Fact~\ref{fac:vector_norm}.

For the first term in Eq.~\eqref{eq:rewrite_H_diff_exp}, we have
\begin{align}\label{eq:upper_bound_H_x_H_y_step_1_exp}
\| A \|^2 \leq R^2
\end{align}

For the second term in Eq.~\eqref{eq:rewrite_H_diff_exp}, we have
\begin{align}\label{eq:upper_bound_H_x_H_y_step_2_exp}
\| ( 2 \exp(Ax) + 2 \exp(Ay) - b ) \|_{\infty} \notag
\leq & ~ \| 2 \exp(Ax)\|_{\infty} + \|2 \exp(Ay)\|_{\infty} + \|b\|_\infty \notag \\
\leq & ~ 2\exp(\|Ax\|_2) + 2\exp(\|Ay\|_2) + \|b\|_\infty \notag \\
\leq & ~ 4 \exp( R^2) + \| b \|_{\infty} \notag \\
\leq & ~ 4 \exp(R^2) + R \notag \\
\leq & ~ 5 \exp(R^2)
\end{align}

where the first step follows from Fact~\ref{fac:vector_norm}
, the second step follows from Fact~\ref{fac:vector_norm}, the third step follows from $\|Ax\|_2 \leq R^2,\|Ay\|_2 \leq R^2$, the fourth step follows from $\|b\|_\infty \leq \|b\|_2 \leq R$, the last step follows from $R \geq 2$.

For the third term in Eq.~\eqref{eq:rewrite_H_diff_exp}, we have
\begin{align}\label{eq:upper_bound_H_x_H_y_step_3_exp}
\| \exp(Ax) - \exp(Ay) \|_2 
\leq & ~ \| \exp(Ax) \|_2 \cdot 2 \| A (y-x) \|_{\infty} \notag \\
\leq & ~ \sqrt{n} \cdot \| \exp(Ax) \|_{\infty}  \cdot 2 \| A (y-x) \|_{\infty} \notag \\
\leq & ~ \sqrt{n} \cdot \| \exp(Ax) \|_{\infty}   \cdot 2 \| A (y-x) \|_2 \notag \\
\leq & ~ \sqrt{n} \cdot \exp(\|Ax\|_2)  \cdot 2 \| A (y-x) \|_2 \notag \\
\leq & ~ \sqrt{n} \exp( R^2)  \cdot 2 \| A \| \cdot \| y - x \|_2 \notag\\
\leq & ~ 2 \sqrt{n} R \exp(R^2) \cdot \|y - x\|_2
\end{align}
where the first step follows  from $\| A (y-x) \|_{\infty} < 0.01$ and Fact~\ref{fac:vector_norm},  
the second step follows from Fact~\ref{fac:vector_norm}, 
the third step follows from Fact~\ref{fac:vector_norm}
, the fourth step follows from Fact~\ref{fac:vector_norm},
the fifth step follows from Fact~\ref{fac:matrix_norm} and $\|Ax\|_2 \leq R^2$, the last step follows from $\|A\| \leq R$.

Putting it all together, we have
\begin{align*}
 \| H(x) - H(y) \| 
 \leq & ~ R^2 \cdot 5 \exp(R^2) \cdot 2 \sqrt{n} R \exp(R^2) \| y - x \|_2 \\
= & ~ 10 \sqrt{n} R^3 \exp(2R^2) \cdot \| y - x \|_2 \\
 \leq & ~ \sqrt{n} \exp(4R^2) \cdot  \exp(2R^2) \cdot \| y - x \|_2 \\
 = & ~ \sqrt{n} \exp(6R^2) \cdot \| y - x \|_2
\end{align*}
where the first step follows from by applying Eq.~\eqref{eq:upper_bound_H_x_H_y_step_1_exp}, Eq.~\eqref{eq:upper_bound_H_x_H_y_step_2_exp}, and Eq.~\eqref{eq:upper_bound_H_x_H_y_step_3_exp}, the second step follows from simple algebra, the third step follows from $R \geq 2$, the last step follows from simple algebra.

\end{proof}

%% file: cosh.tex
\section{Cosh Regression}\label{sec:cosh}
 
In this section, we provide detailed analysis of $L_{\cosh}$. In Section~\ref{sec:cosh:definition} we define the loss function $L_{\cosh}$ based on $\cosh(x)$. In Section~\ref{sec:cosh:gradient} we compute the gradient of $L_{\cosh}$ by detail. In Section~\ref{sec:cosh:hessian} we compute the hessian of $L_{\cosh}$ by detail. In Section~\ref{sec:cosh:gradient_hessian}, we summarize the result of Section~\ref{sec:cosh:gradient} and Section~\ref{sec:cosh:hessian} and aquire the gradient $\nabla L_{\cosh}$ and hessian $\nabla^2 L_{\cosh}$ for $L_{\cosh}$. In Section~\ref{sec:cosh:loss_and_reg} we define $L_{\cosh,\reg}$ by adding the regularization term $L_{\reg}$ in Section~\ref{sec:preli:regularization} to $L_{\cosh}$ and compute the gradient $\nabla L_{\cosh,\reg}$ and hessian $\nabla^2 L_{\cosh,\reg}$ of $L_{\cosh,\reg}$. In Section~\ref{sec:cosh:convex} we proved that $\nabla^2 L_{\cosh,\reg} \succ 0$ and thus showed that $L_{\cosh,\reg}$ is convex. In Section~\ref{sec:cosh:lipschitz} we provide the upper bound for $\|\nabla^2 L_{\cosh,\reg}(x)-\nabla^2 L_{\cosh,\reg}(y)\|$ and thus proved $\nabla^2 L_{\cosh,\reg}$ is lipschitz.
\subsection{Definition}\label{sec:cosh:definition}

\begin{definition}\label{def:L_cosh}
Given $A \in \R^{n \times d}$ and $b \in \R^n$. For a vector $x \in \R^d$, we define loss function $L_{\cosh}(x)$ as follows:
\begin{align*}
L_{\cosh}(x) := 0.5 \cdot \| \cosh(Ax) - b \|_2^2
\end{align*}
\end{definition}

\subsection{Gradient}\label{sec:cosh:gradient}

\begin{lemma}[Gradient for Cosh]\label{lem:gradient_cosh}
We have
\begin{itemize}
    
    \item Part 1.
    \begin{align*}
        \frac{ \d ( \cosh(Ax) - b ) }{ \d t } = \sinh(Ax) \circ \frac{A \d x }{ \d t}
    \end{align*}
    \item Part 2. 
    \begin{align*}
        \frac{\d L_{\cosh} }{ \d t} = (\cosh(Ax) - b )^\top \cdot ( \sinh(Ax) \circ \frac{A \d x}{ \d t } )
    \end{align*}
\end{itemize}
Further, we have for each $i \in [d]$
\begin{itemize}
    
    \item Part 3.
    \begin{align*}
        \frac{ \d ( \cosh(Ax) - b ) }{ \d x_i } = \sinh(Ax) \circ A_{*,i}
    \end{align*}
    \item Part 4. 
    \begin{align*}
        \frac{\d L_{\cosh} }{ \d x_i}  
        = & ~ (\cosh(Ax) - b )^\top \cdot ( \sinh(Ax) \circ A_{*,i} )
    \end{align*}
    \item Part 5.
    \begin{align*}
        \frac{\d L_{\cosh} }{ \d x} =  A^\top \diag( \sinh(Ax) ) (\cosh(Ax) - b)
    \end{align*}
\end{itemize}
\end{lemma}

\begin{proof}

{\bf Proof of Part 1.}

For each $i \in [n]$, we have
\begin{align*}
 \frac{ \d ( \cosh(Ax) - b )_i }{ \d t } 
 = & ~ \sinh(Ax)_i \cdot \frac{\d (Ax)_i}{\d t} \\
 = & ~ \sinh(Ax)_i \cdot \frac{ (A \d x)_i}{\d t} 
\end{align*}
where the first and second step follow from the differential chain rule.

Thus, we complete the proof.

{\bf Proof of Part 2.}

We have
 
\begin{align*}
    \frac{\d L_{\cosh} }{ \d t} 
    = & ~ (\cosh(Ax) - b )^\top \cdot \frac{ \d ( \cosh(Ax) - b ) }{\d t} \\
    = & ~ (\cosh(Ax) - b )^\top \cdot ( \sinh(Ax) \circ \frac{A \d x}{ \d t } )
\end{align*}
where the first step follows from the differential chain rule, and the second step follows from $\frac{ \d ( \cosh(Ax) - b ) }{ \d t } = \sinh(Ax) \circ \frac{A \d x }{ \d t}$ in {\bf Part 1}.

{\bf Proof of Part 3.}

We have
\begin{align*}
    \frac{\d (\cosh(Ax)-b)}{\d x_i}
    = & ~ \frac{\d (\cosh(Ax))}{\d x_i} - \frac{\d b}{\d x_i}\\
    = & ~ \sinh(Ax) \circ \frac{\d Ax}{\d x_i} - 0\\
    = & ~ \sinh(Ax) \circ A_{*,i}
\end{align*}
where the first step follows from the property of the gradient, the second step follows from simple algebra, and the last step directly follows from Lemma~\ref{lem:Ax_gradient_hessian}.

{\bf Proof of Part 4.}

By substitute $x_i$ into $t$ of {\bf Part 3}, we get
\begin{align*}
    \frac{\d L}{\d x_i}
    = & ~ (\cosh(Ax) - b )^\top \cdot ( \sinh(Ax) \circ \frac{A \d x}{ \d x_i } )\\
    = & ~ (\cosh(Ax) - b )^\top \cdot ( \sinh(Ax) \circ A_{*,i} ) \\
    = & ~ A_{*,i}^\top \diag( \sinh(Ax) ) (\cosh(Ax) - b )
\end{align*}
where the first step follows from the result of {\bf Part 2},
the second step follows from the result of Lemma~\ref{lem:Ax_gradient_hessian},
the last step follows from Fact~\ref{fac:circ_diag}.

{\bf Proof of Part 5.}

We have
\begin{align*}
    \frac{\d L}{ \d x}
    = & ~ A^\top \diag( \sinh(Ax) ) (\cosh(Ax) - b)
\end{align*}
this follows from the result of {\bf Part 4} directly.
\end{proof}

\subsection{Hessian}\label{sec:cosh:hessian}

\begin{lemma}\label{lem:hessian_cosh}
\begin{itemize}

    \item Part 1.
    \begin{align*}
        \frac{ \d^2 ( \cosh(Ax) - b ) }{ \d x_i^2 }
        = & ~ A_{*,i} \circ \cosh(Ax) \circ A_{*,i}
    \end{align*}
    \item Part 2.
    \begin{align*}
        \frac{ \d^2 ( \cosh(Ax) - b ) }{ \d x_i \d x_j }
        = & ~ A_{*,j} \circ \cosh(Ax) \circ A_{*,i}
    \end{align*}
    \item Part 3. 
    \begin{align*}
        \frac{\d^2 L_{\cosh} }{ \d x_i^2}
        = & ~ A_{*,i}^\top \diag( 2 \cosh(Ax) \circ \cosh(Ax) - b \circ \cosh(Ax) - {\bf 1}_n )  A_{*,i}
    \end{align*}
    \item Part 4. 
     \begin{align*}
        \frac{\d^2 L_{\cosh} }{ \d x_i \d x_j} = A_{*,i}^\top \diag( \sinh^2(Ax) + \cosh^2(Ax) - b \circ \cosh(Ax) ) A_{*,j}
    \end{align*}
\end{itemize}
\end{lemma}
\begin{proof}

{\bf Proof of Part 1.}
\begin{align*}
    \frac{ \d^2 ( \cosh(Ax) - b ) }{ \d x_i^2 }
        = & ~ \frac{\d}{\d x_i}\bigg(\frac{\d (\cosh(Ax) - b)}{\d x_i} \bigg) \\
        = & ~ \frac{\d (\sinh(Ax) \circ A_{*,i})}{\d x_i} \\
        = & ~ A_{*,i} \circ \frac{\d \sinh(Ax)}{\d x_i} \\
        = & ~ A_{*,i} \circ \cosh(Ax) \circ A_{*,i}
\end{align*}
where the first step is an expansion of the Hessian, 
the second step follows from the differential chain rule, 
the third step extracts the matrix $A_{*,i}$ with constant entries out of the derivative, 
and the last step also follows from the chain rule.

{\bf Proof of Part 2.}
\begin{align*}
    \frac{ \d^2 ( \cosh(Ax) - b ) }{ \d x_i \d x_j }
        = & ~ \frac{\d}{\d x_i}\bigg(\frac{\d}{\d x_j}\bigg(\cosh(Ax) - b\bigg) \bigg) \\
        = & ~ \frac{\d}{\d x_i}\bigg(\sinh(Ax) \circ A_{*,j} \bigg) \\
        = & ~ A_{*,j} \circ \cosh(Ax) \circ A_{*,i}
\end{align*}
where the first step is an expansion of the Hessian, the second and third steps follow from the differential chain rule.

{\bf Proof of Part 3.}

Here in the proof, for simplicity, we let use $\cosh^2(Ax) = \cosh(Ax) \circ \cosh(Ax)$.

We have
\begin{align*}
    \frac{\d^2 L }{ \d x_i^2}
        = & ~ \frac{\d}{\d x_i}\bigg(\frac{\d L}{\d x_i} \bigg) \\
        = & ~ \frac{\d}{\d x_i}\bigg((\cosh(Ax) - b )^\top \cdot ( \sinh(Ax) \circ A_{*,i} ) \bigg) \\
        = & ~ (\sinh(Ax) \circ A_{*,i})^\top \cdot ( \sinh(Ax) \circ A_{*,i} )+(\cosh(Ax) - b )^\top \cdot (A_{*,i} \circ \cosh(Ax) \circ A_{*,i}) \\ 
        = & ~ A_{*,i}^\top \diag( \sinh(Ax) \circ \sinh(Ax) + \cosh(Ax) \circ \cosh(Ax) - b \circ \cosh(Ax) ) A_{*,i} \\
        = & ~ A_{*,i}^\top \diag( 2 \cosh(Ax) \circ \cosh(Ax) - b \circ \cosh(Ax) - {\bf 1}_n ) A_{*,i}
\end{align*}
where the first step is an expansion of the Hessian, 
the second step follows from {\bf Part 4} of Lemma~\ref{lem:gradient_cosh},
the third step follows from the product rule of calculus,
the fourth step follows from Fact~\ref{fac:circ_diag},
the last step follows from Fact~\ref{fac:circ_diag}.

{\bf Proof of Part 4.}
\begin{align*}
    \frac{\d^2 L }{ \d x_i \d x_j}
        = & ~ \frac{\d}{\d x_i}\bigg(\frac{\d L}{\d x_j} \bigg) \\
        = & ~ \frac{\d}{\d x_i}\bigg((\cosh(Ax) - b )^\top \cdot ( \sinh(Ax) \circ A_{*,j} ) \bigg) \\
        = & ~ (\sinh(Ax) \circ A_{*,i})^\top \cdot ( \sinh(Ax) \circ A_{*,j} )+(\cosh(Ax) - b )^\top \cdot (A_{*,j} \circ \cosh(Ax) \circ A_{*,i}) \\
        = & ~ A_{*,i}^\top \diag( \sinh(Ax) \circ \sinh(Ax) + \cosh(Ax) \circ \cosh(Ax) - b \circ \cosh(Ax) ) A_{*,j} \\
        = & ~ A_{*,i}^\top \diag( 2 \cosh(Ax) \circ \cosh(Ax) - b \circ \cosh(Ax) - {\bf 1}_n ) A_{*,j}
\end{align*}
where the first step is an expansion of the Hessian, 
the second step follows from {\bf Part 4} of Lemma~\ref{lem:gradient_cosh},
the third step follows from the product rule of calculus,
the fourth step follows from Fact~\ref{fac:circ_diag},
the last step follows from Fact~\ref{fac:circ_diag}.

\end{proof}

\subsection{Gradient and Hessian of the Loss function for Cosh Function}\label{sec:cosh:gradient_hessian}

\begin{lemma}\label{lem:gradient_hessian_cosh}
    Let $L: \R^d \to \R_{\geq 0}$ be defined in Definition~\ref{def:L_exp}. Then for any $i, j \in [d]$, we have
    \begin{itemize}
        \item Part 1. Gradient
    \begin{align*}
        \nabla L_{\cosh} = A^\top \diag(\sinh(Ax)) \diag(\cosh(Ax) - b) {\bf 1}_n
    \end{align*}
        \item Part 2. Hessian
        \begin{align*}
            \nabla^2 L_{\cosh} = A^\top \diag( 2 \cosh(Ax) \circ \cosh(Ax) - b \circ \cosh(Ax) ) A
        \end{align*}
    \end{itemize}
\end{lemma}
 
\begin{proof}

{\bf Part 1.}
We run Lemma~\ref{lem:gradient_cosh} and Fact~\ref{fac:circ_diag} directly.

{\bf Part 2.}
It follows from Part 5 of Lemma~\ref{lem:hessian_cosh}.
\end{proof}

\subsection{Loss Function with a Regularization Term}\label{sec:cosh:loss_and_reg}

\begin{definition}\label{def:L_cosh_and_regularized}
Given matrix $A \in \R^{n \times d}$ and $b \in \R^n$, $w \in \R^n$. For a vector $x \in \R^d$, we define loss function $L(x)$ as follows
\begin{align*}
L_{\cosh,\reg}(x): = 0.5 \cdot \| \cosh(Ax) - b \|_2^2+ 0.5 \cdot \| W A x \|_2^2
\end{align*}
where $W = \diag(w)$.
\end{definition}

\begin{lemma}
Let $L$ be defined as Definition~\ref{def:L_cosh_and_regularized}, then we have
\begin{itemize}
    \item Part 1. Gradient
    \begin{align*}
        \frac{\d L_{\cosh,\reg}}{\d x} = A^\top \diag(\sinh(Ax)) ( \diag(\cosh(Ax) - b) ) {\bf 1}_n + A^\top W^2 A x
    \end{align*}
    \item Part 2. Hessian
    \begin{align*} 
        \frac{\d^2 L_{\cosh,\reg}}{\d x^2} = A^\top \diag( 2  \cosh(Ax) \circ \cosh(Ax) - b \circ \cosh(Ax) ) A +  A^\top W^2 A
    \end{align*}
    
\end{itemize}
\end{lemma}
\begin{proof}
{\bf Proof of Part 1.}
We run Lemma~\ref{lem:gradient_hessian_exp} and Lemma~\ref{lem:regularization} directly.

{\bf Proof of Part 2.}
We run Lemma~\ref{lem:gradient_hessian_exp} and Lemma~\ref{lem:regularization} directly.

\end{proof}

\subsection{Hessian is Positive Definite}\label{sec:cosh:convex}

\begin{lemma}\label{lem:hessian_is_pd_cosh}
 If $w_{i}^2 > 0.5 b_{i}^2 + l/\sigma_{\min}(A)^2 + 1$ for all $i \in [n]$, then 
    \begin{align*}
        \frac{\d^2 L}{\d x^2} \succeq l \cdot I_d
    \end{align*}
\end{lemma}

\begin{proof}
We define diagonal matrix $D \in \R^{n \times n}$
\begin{align*}
D = \diag( \sinh^2(Ax) + \cosh^2(Ax) - b \circ \cosh(Ax) ) + W^2
\end{align*}

Then we can rewrite Hessian as 
\begin{align*}
\frac{\d^2 L}{\d x^2} = A^\top D A.
\end{align*}

Then we have
\begin{align*}
D_{i,i} 
= & ~ ( \sinh^2( (Ax)_i ) + \cosh^2((Ax)_i) - b_i \cosh^2((Ax)_i) ) ) + w_{i,i}^2 \\
= & ~ ( z_i^2 - 1 + z_i^2 - b_i z_i ) + w_{i}^2 \\
= & ~ 2 z_i^2 - b_i z_i + w_i^2 -1 \\
> & ~ 2 z_i^2 - b_i z_i + 0.5 b_{i}^2 + l/\sigma_{\min}(A)^2 \\
= & ~ 0.5 ( 2z_i - b_i )^2 + l/\sigma_{\min}(A)^2 \\
\geq & ~ l/\sigma_{\min}(A)^2
\end{align*}
where the first step follows from simple algebra, the second step follows from replacing $\cosh(Ax)$ with $z = \cosh(Ax)$ and $\sinh^2() = \cosh^2()-1$ (Fact~\ref{fac:e_cosh_sinh_exact}), the third step follows from $w_{i}^2 > 0.5b_{i}^2 + l/\sigma_{\min}(A)^2 + 1$, the fourth step follows from simple algebra, the fifth step follows from $x^2\geq0, \forall x$.

Since we know $D_{i,i} > 0$ for all $i \in [n]$ and Lemma~\ref{lem:ADA_pd}, we have 
\begin{align*}
A^\top D A \succeq (\min_{i \in [n]} D_{i,i}) \cdot \sigma_{\min}(A)^2 I_d \succeq l \cdot I_d
\end{align*}
Thus, Hessian is positive definite forever and thus the function is convex.
\end{proof}

\subsection{Hessian is Lipschitz}\label{sec:cosh:lipschitz}
\begin{lemma}[Hessian is Lipschitz]\label{lem:hessian_is_lipschitz_cosh}
If the following condition holds
 
\begin{itemize}
    \item Let $H(x) = \frac{\d^2 L_{\cosh,\reg}}{\d x^2}$
    \item Let $R > 2$
    \item $\|x \|_2 \leq R, \| y \|_2 \leq R$
    \item $\| A (x-y) \|_{\infty} < 0.01$
    \item $\| A \| \leq R$
    \item $\| b \|_2 \leq R$
\end{itemize}
Then we have
    \begin{align*}
        \| H(x) - H(y) \| \leq \sqrt{n} \exp(6R^2) \cdot \| x- y \|_2
    \end{align*}
\end{lemma}
\begin{proof}

We have
\begin{align}\label{eq:rewrite_H_diff_cosh}
& ~ \| H(x) - H(y) \| \notag \\
= & ~ \| A^\top \diag(2 \cosh(Ax) - b) \diag(\cosh(Ax)) A -  A^\top \diag(2 \cosh(Ay) - b) \diag(\cosh(Ay)) A \| \notag \\
\leq & ~ \| A \|^2 \cdot \|  (2 \cosh(Ax) - b) \circ \cosh(Ax) - (2 \cosh(Ay) - b) \circ \cosh(Ay) \|_2  \notag \\
= & ~ \| A \|^2 \cdot \| 2 (\cosh(Ax) + \cosh(Ay) )\circ ( \cosh(Ax) - \cosh(Ay) ) - b \circ ( \cosh(Ax) - \cosh(A y) ) \|_2 \notag \\
= & ~ \| A \|^2 \cdot \| ( 2 \cosh(Ax) + 2 \cosh(Ay) - b ) \circ ( \cosh(Ax) - \cosh(Ay) ) \|_2  \notag \\
\leq & ~ \| A \|^2 \cdot \| ( 2 \cosh(Ax) + 2 \cosh(Ay) - b ) \|_{\infty} \cdot \| \cosh(Ax) - \cosh(Ay) \|_2
\end{align}
where the first step follows from $H(x) = \nabla^2L$ and simple algebra, 
the second step follows from Fact~\ref{fac:matrix_norm}, the third step follows from simple algebra, the fourth step follows from simple algebra, the last step follows from Fact~\ref{fac:vector_norm}.

For the first term in Eq.~\eqref{eq:rewrite_H_diff_cosh}, we have
\begin{align}\label{eq:upper_bound_H_x_H_y_step_1_cosh}
\| A \|^2 \leq R^2
\end{align}

For the second term in Eq.~\eqref{eq:rewrite_H_diff_cosh}, we have
\begin{align}\label{eq:upper_bound_H_x_H_y_step_2_cosh}
\| ( 2 \cosh(Ax) + 2 \cosh(Ay) - b ) \|_{\infty} \notag
\leq & ~ \| 2 \cosh(Ax)\|_{\infty} + \|2 \cosh(Ay)\|_{\infty} + \|b\|_\infty \notag \\
\leq & ~ 2\exp(\|Ax\|_2) + 2\exp(\|Ay\|_2) + \|b\|_\infty \notag \\
\leq & ~ 4 \exp(R^2) + \| b \|_{\infty} \notag \\
\leq & ~ 4 \exp(R^2) + R \notag \\
\leq & ~ 5 \exp(R^2)
\end{align}

where the first step follows from Fact~\ref{fac:vector_norm}
, 
the second step follows from Fact~\ref{fac:vector_norm}
, 
the third step follows from $\|Ax\|_2 \leq R^2$, $\|Ay\|_2 \leq R^2$, the fourth step follows from $\|b\|_\infty \leq R$ and the last step follows from $R \geq 2$.

For the third term in Eq.~\eqref{eq:rewrite_H_diff_cosh}, we have
\begin{align}\label{eq:upper_bound_H_x_H_y_step_3_cosh}
\| \cosh(Ax) - \cosh(Ay) \|_2 
\leq & ~ \| \cosh(Ax) \|_2 \cdot 2 \| A (y-x) \|_{\infty} \notag \\
\leq & ~ \sqrt{n}\| \cosh(Ax) \|_\infty \cdot 2 \| A (y-x) \|_{\infty} \notag \\
\leq & ~ \sqrt{n}\exp(\|Ax\|_2) \cdot 2 \| A (y-x) \|_{2} \notag \\
\leq & ~ \sqrt{n}\exp(R^2)  \cdot 2 \| A (y-x) \|_2 \notag \\
\leq & ~ \sqrt{n}\exp(R^2)  \cdot 2 \| A \| \cdot \| y - x \|_2 \notag\\
\leq & ~ 2 \sqrt{n}R \exp(R^2) \cdot \|y - x\|_2
\end{align}
where the first step follows  from $\| A (y-x) \|_{\infty} < 0.01$ and Fact~\ref{fac:vector_norm}, the second step follows from Fact~\ref{fac:vector_norm}, the third step follows from Fact~\ref{fac:vector_norm},   the fourth step follows from $\|Ax\|_2 \leq R^2$, the fifth step follows from Fact~\ref{fac:matrix_norm}, the last step follows from $\|A\| \leq R$.

Putting it all together, we have
\begin{align*}
 \| H(x) - H(y) \| 
 \leq & ~ R^2 \cdot 5 \exp(R^2) \cdot 2\sqrt{n} R\exp(R^2) \| y - x \|_2 \\
= & ~ 10\sqrt{n}  R^3 \exp(2R^2) \cdot \| y - x \|_2 \\
 \leq & ~ \sqrt{n} \exp(4R^2) \cdot  \exp(2R^2) \cdot \| y - x \|_2 \\
 = & ~ \sqrt{n}  \exp(6R^2) \cdot \| y - x \|_2
\end{align*}
where the first step follows from by applying Eq.~\eqref{eq:upper_bound_H_x_H_y_step_1_cosh}, Eq.~\eqref{eq:upper_bound_H_x_H_y_step_2_cosh}, and Eq.~\eqref{eq:upper_bound_H_x_H_y_step_3_cosh}, the second step follows from simple algebra, the third step follows from $R \geq 2$, the last step follows from simple algebra.

\end{proof}

%% file: sinh.tex
\section{Sinh Regression}\label{sec:sinh}
 
In this section, we provide detailed analysis of $L_{\sinh}$. In Section~\ref{sec:sinh:definition} we define the loss function $L_{\sinh}$ based on $\sinh(x)$. In Section~\ref{sec:sinh:gradient} we compute the gradient of $L_{\sinh}$ by detail. In Section~\ref{sec:sinh:hessian} we compute the hessian of $L_{\sinh}$ by detail. In Section~\ref{sec:sinh:gradient_hessian}, we summarize the result of Section~\ref{sec:sinh:gradient} and Section~\ref{sec:sinh:hessian} and aquire the gradient $\nabla L_{\sinh}$ and hessian $\nabla^2 L_{\sinh}$ for $L_{\sinh}$. In Section~\ref{sec:sinh:loss_reg} we define $L_{\sinh,\reg}$ by adding the regularization term 
$L_{\reg}$ in Section~\ref{sec:preli:regularization} to $L_{\sinh}$ and compute the gradient $\nabla L_{\sinh,\reg}$ and hessian $\nabla^2 L_{\sinh,\reg}$ of $L_{\sinh,\reg}$. In Section~\ref{sec:sinh:convex} we proved that $\nabla^2 L_{\sinh,\reg} \succ 0$ and thus showed that $L_{\sinh,\reg}$ is convex. In Section~\ref{sec:sinh:lipschitz} we provide the upper bound for $\|\nabla^2 L_{\sinh,\reg}(x)-\nabla^2 L_{\sinh,\reg}(y)\|$ and thus proved $\nabla^2 L_{\sinh,\reg}$ is lipschitz.
\subsection{Definition}\label{sec:sinh:definition}
\begin{definition}\label{def:L_sinh}
Given $A \in \R^{n \times d}$ and $b \in \R^n$. For a vector $x \in \R^d$, we define loss function $L(x)$ as follows:
\begin{align*}
L(x) := 0.5 \cdot \| \sinh(Ax) - b \|_2^2
\end{align*}
\end{definition}

\subsection{Gradient}\label{sec:sinh:gradient}

\begin{lemma}[Gradient for Sinh]\label{lem:gradient_sinh}
We have
\begin{itemize}
    
    \item Part 1.
    \begin{align*}
        \frac{ \d ( \sinh(Ax) - b ) }{ \d t } = \cosh(Ax) \circ \frac{A \d x }{ \d t}
    \end{align*}
    \item Part 2. 
    \begin{align*}
        \frac{\d L_{\sinh} }{ \d t} = (\sinh(Ax) - b )^\top \cdot ( \cosh(Ax) \circ \frac{A \d x}{ \d t } )
    \end{align*}
\end{itemize}
Further, we have for each $i \in [d]$
\begin{itemize}
    
    \item Part 3.
    \begin{align*}
        \frac{ \d ( \sinh(Ax) - b ) }{ \d x_i } = \cosh(Ax) \circ A_{*,i}
    \end{align*}
    \item Part 4. 
    \begin{align*}
        \frac{\d L_{\sinh} }{ \d x_i}  
        = & ~ (\sinh(Ax) - b )^\top \cdot ( \cosh(Ax) \circ A_{*,i} )
    \end{align*}
    \item Part 5.
    \begin{align*}
        \frac{\d L_{\sinh} }{ \d x} =  A^\top \diag( \cosh(Ax) ) (\sinh(Ax) - b)
    \end{align*}
\end{itemize}
\end{lemma}

\begin{proof}

{\bf Proof of Part 1.}

For each $i \in [n]$, we have
\begin{align*}
 \frac{ \d ( \sinh(Ax) - b )_i }{ \d t } 
 = & ~ \cosh(Ax)_i \cdot \frac{\d (Ax)_i}{\d t} \\
 = & ~ \cosh(Ax)_i \cdot \frac{ (A \d x)_i}{\d t} 
\end{align*}
where the first and second step follow from the differential chain rule.

Thus, we complete the proof.

{\bf Proof of Part 2.}

We have
 
\begin{align*}
    \frac{\d L_{\sinh} }{ \d t} 
    = & ~ (\sinh(Ax) - b )^\top \cdot \frac{ \d ( \sinh(Ax) - b ) }{\d t} \\
    = & ~ (\sinh(Ax) - b )^\top \cdot ( \cosh(Ax) \circ \frac{A \d x}{ \d t } )
\end{align*}
where the first step follows from the differential chain rule, and the second step follows from $\frac{ \d ( \sinh(Ax) - b ) }{ \d t } = \cosh(Ax) \circ \frac{A \d x }{ \d t}$ in {\bf Part 1}.

{\bf Proof of Part 3.}

We have
\begin{align*}
    \frac{\d (\sinh(Ax)-b)}{\d x_i}
    = & ~ \frac{\d (\sinh(Ax))}{\d x_i} - \frac{\d b}{\d x_i}\\
    = & ~ \cosh(Ax) \circ \frac{\d Ax}{\d x_i} - 0\\
    = & ~ \cosh(Ax) \circ A_{*,i}
\end{align*}
where the first step follows from the property of the gradient, the second step follows from the differential chain rule, and the last step directly follows from Lemma~\ref{lem:Ax_gradient_hessian}.

{\bf Proof of Part 4.}

By substitute $x_i$ into $t$ of {\bf Part 3}, we get
\begin{align*}
    \frac{\d L}{\d x_i}
    = & ~ (\sinh(Ax) - b )^\top \cdot ( \cosh(Ax) \circ \frac{A \d x}{ \d x_i } )\\
    = & ~ (\sinh(Ax) - b )^\top \cdot ( \cosh(Ax) \circ A_{*,i} ) \\
    = & ~ A_{*,i}^\top \diag( \cosh(Ax) ) (\sinh(Ax) - b )
\end{align*}
where the first step follows from the result of {\bf Part 2} and the second step follows from the result of Lemma~\ref{lem:Ax_gradient_hessian}, the last step follows from Fact~\ref{fac:circ_diag}.

{\bf Proof of Part 5.}

We have
\begin{align*}
    \frac{\d L}{ \d x}
    = & ~ A^\top \diag( \cosh(Ax) ) (\sinh(Ax) - b)
\end{align*}
where this step follows from the result of {\bf Part 4} directly.
\end{proof}

\subsection{Hessian}\label{sec:sinh:hessian}

\begin{lemma}\label{lem:hessian_sinh}
\begin{itemize}

    \item Part 1.
    \begin{align*}
        \frac{ \d^2 ( \sinh(Ax) - b ) }{ \d x_i^2 }
        = & ~ A_{*,i} \circ \sinh(Ax) \circ A_{*,i}
    \end{align*}
    \item Part 2.
    \begin{align*}
        \frac{ \d^2 ( \sinh(Ax) - b ) }{ \d x_i \d x_j }
        = & ~ A_{*,j} \circ \sinh(Ax) \circ A_{*,i}
    \end{align*}
    \item Part 3.
    \begin{align*}
        \frac{\d^2 L_{\sinh} }{ \d x_i^2}
        = & ~ A_{*,i}^\top \diag( 2\sinh(Ax) \circ \sinh(Ax) - b \circ \sinh(Ax) + {\bf 1}_n )  A_{*,i}
    \end{align*}
    \item Part 4. 
     \begin{align*}
        \frac{\d^2 L_{\sinh} }{ \d x_i \d x_j} = A_{*,i}^\top \diag( 2 \sinh(Ax) \circ \sinh(Ax) - b \circ \sinh(Ax) - {\bf 1}_n ) A_{*,j}
    \end{align*}
\end{itemize}
\end{lemma}
\begin{proof}

{\bf Proof of Part 1.}
\begin{align*}
    \frac{ \d^2 ( \sinh(Ax) - b ) }{ \d x_i^2 }
        = & ~ \frac{\d}{\d x_i}\bigg(\frac{\d (\sinh(Ax) - b)}{\d x_i} \bigg) \\
        = & ~ \frac{\d (\cosh(Ax) \circ A_{*,i})}{\d x_i} \\
        = & ~ A_{*,i} \circ \frac{\d \cosh(Ax)}{\d x_i} \\
        = & ~ A_{*,i} \circ \sinh(Ax) \circ A_{*,i}
\end{align*}
where the first step is an expansion of the Hessian, the second step follows from the differential chain rule, the third step extracts the matrix $A_{*,i}$ with constant entries out of the derivative, and the last step also follows from the chain rule.

{\bf Proof of Part 2.}
\begin{align*}
    \frac{ \d^2 ( \sinh(Ax) - b ) }{ \d x_i \d x_j }
        = & ~ \frac{\d}{\d x_i}\bigg(\frac{\d}{\d x_j}\bigg(\sinh(Ax) - b\bigg) \bigg) \\
        = & ~ \frac{\d}{\d x_i}\bigg(\cosh(Ax) \circ A_{*,j} \bigg) \\
        = & ~ A_{*,j} \circ \sinh(Ax) \circ A_{*,i}
\end{align*}
where the first step is an expansion of the Hessian, the second and third steps follow from the differential chain rule.

{\bf Proof of Part 3.}
\begin{align*}
    \frac{\d^2 L }{ \d x_i^2}
        = & ~ \frac{\d}{\d x_i}\bigg(\frac{\d L}{\d x_i} \bigg) \\
        = & ~ \frac{\d}{\d x_i}\bigg((\sinh(Ax) - b )^\top \cdot ( \cosh(Ax) \circ A_{*,i} ) \bigg) \\
        = & ~ (\cosh(Ax) \circ A_{*,i})^\top \cdot ( \cosh(Ax) \circ A_{*,i} )+(\sinh(Ax) - b )^\top \cdot (A_{*,i} \circ \sinh(Ax) \circ A_{*,i})\\ 
        = & ~ A_{*,i}^\top \diag( \cosh^2(Ax) + \sinh^2(Ax) - b \circ \sinh(Ax) ) A_{*,i} \\
        = & ~ A_{*,i}^\top \diag( 2\sinh(Ax) \circ \sinh(Ax) - b \circ \sinh(Ax) + {\bf 1}_n )  A_{*,i}
\end{align*}
where the first step is an expansion of the Hessian,
the second step follows from {\bf Part 4} of Lemma~\ref{lem:gradient_sinh},
the third step follows from the product rule of calculus,
the fourth step follows from Fact~\ref{fac:circ_diag},
the last step follows from Fact~\ref{fac:circ_diag}.

{\bf Proof of Part 4.}
\begin{align*}
    \frac{\d^2 L }{ \d x_i \d x_j}
        = & ~ \frac{\d}{\d x_i}\bigg(\frac{\d L}{\d x_j} \bigg) \\
        = & ~ \frac{\d}{\d x_i}\bigg((\sinh(Ax) - b )^\top \cdot ( \cosh(Ax) \circ A_{*,j} ) \bigg) \\
        = & ~ (\cosh(Ax) \circ A_{*,i})^\top \cdot ( \cosh(Ax) \circ A_{*,j} )+(\sinh(Ax) - b )^\top \cdot (A_{*,j} \circ \sinh(Ax) \circ A_{*,i}) \\
        = & ~ A_{*,i}^\top \diag( \cosh^2(Ax) + \sinh^2(Ax) - b \circ \sinh(Ax) ) A_{*,j} \\
        = & ~ A_{*,i}^\top \diag( 2\sinh(Ax) \circ \sinh(Ax) - b \circ \sinh(Ax) + {\bf 1}_n )  A_{*,j}
\end{align*}
where the first step is an expansion of the Hessian,
the second step follows from {\bf Part 4} of Lemma~\ref{lem:gradient_sinh},
the third step follows from the product rule of calculus,
the fourth step follows from Fact~\ref{fac:circ_diag},
the last step follows from Fact~\ref{fac:circ_diag}.

\end{proof}

\subsection{Gradient and Hessian of the Loss function for Sinh Function}\label{sec:sinh:gradient_hessian}

\begin{lemma}\label{lem:gradient_hessian_sinh}
    Let $L: \R^d \to \R_{\geq 0}$ be defined in Definition~\ref{def:L_exp}. Then for any $i, j \in [d]$, we have
    \begin{itemize}
        \item Part 1. Gradient
    \begin{align*}
        \nabla L_{\sinh} = A^\top \diag(\cosh(Ax)) \diag(\sinh(Ax) - b) {\bf 1}_n
    \end{align*}
        \item Part 2. Hessian
        \begin{align*}
            \nabla^2 L_{\sinh} = A^\top \diag( 2\sinh(Ax) \circ \sinh(Ax) - b \circ \sinh(Ax) + {\bf 1}_n ) A
        \end{align*}
    \end{itemize}
\end{lemma}
 
\begin{proof}

{\bf Part 1.}
We run Lemma~\ref{lem:gradient_sinh} and Fact~\ref{fac:circ_diag} directly.

{\bf Part 2.}
It follows from Part 5 of Lemma~\ref{lem:hessian_sinh}.
\end{proof}

\subsection{Loss Function with a Regularization Term}\label{sec:sinh:loss_reg}

\begin{definition}\label{def:L_sinh_and_regularized}
Given matrix $A \in \R^{n \times d}$ and $b \in \R^n$, $w \in \R^n$. For a vector $x \in \R^d$, we define loss function $L(x)$ as follows
\begin{align*}
L_{\sinh,\reg}(x): = 0.5 \cdot \| \sinh(Ax) - b \|_2^2+ 0.5 \cdot \| W A x \|_2^2
\end{align*}
where $W = \diag(w)$.
\end{definition}

\begin{lemma}
Let $L$ be defined as Definition~\ref{def:L_sinh_and_regularized}, then we have
\begin{itemize}
    \item Part 1. Gradient
    \begin{align*}
        \frac{\d L_{\sinh,\reg}}{\d x} = A^\top \diag(\cosh(Ax)) ( \diag(\sinh(Ax) - b) ) {\bf 1}_n + A^\top W^2 A x
    \end{align*}
    \item Part 2. Hessian
    \begin{align*} 
        \frac{\d^2 L_{\sinh,\reg}}{\d x^2} = A^\top \diag( 2\sinh(Ax) \circ \sinh(Ax) - b \circ \sinh(Ax) + {\bf 1}_n ) A +  A^\top W^2 A
    \end{align*}
\end{itemize}
\end{lemma}
\begin{proof}
{\bf Proof of Part 1.}
We run Lemma~\ref{lem:gradient_hessian_exp} and Lemma~\ref{lem:regularization} directly.

{\bf Proof of Part 2.}
We run Lemma~\ref{lem:gradient_hessian_exp} and Lemma~\ref{lem:regularization} directly.
\end{proof}

\subsection{Hessian is Positive Definite}\label{sec:sinh:convex}
\begin{lemma}\label{lem:hessian_is_pd_sinh}
Let $l > 0$ denote a parameter. 
 If $w_{i}^2 > 0.5 b_{i}^2 + l/\sigma_{\min}(A)^2 - 1$ for all $i \in [n]$, then
    \begin{align*}
        \frac{\d^2 L}{\d x^2} \succeq l \cdot I_d
    \end{align*}
\end{lemma}

\begin{proof}

We define $D$
\begin{align*}
D = \diag( 2\sinh(Ax) \circ \sinh(Ax) - b \circ \sinh(Ax) + {\bf 1}_n ) + W^2
\end{align*}

Then we can rewrite Hessian as 
\begin{align*}
\frac{\d^2 L}{\d x^2} = A^\top D A.
\end{align*} 

We define
\begin{align*}
z_i = \sinh( ( Ax )_i )
\end{align*}
 
Then we have
\begin{align*}
D_{i,i} 
= & ~ ( 2 \sinh^2((Ax)_i) +1 - b_i \sinh((Ax)_i) ) ) + w_{i,i}^2 \\
= & ~ ( 2z_i^2 + 1  - b_i z_i ) + w_{i}^2 \\
= & ~ 2 z_i^2 - b_i z_i + w_i^2 + 1 \\
> & ~ 2 z_i^2 - b_i z_i + 0.5 b_{i}^2 + l/\sigma_{\min}(A)^2\\
= & ~ 0.5 ( 2z_i - b_i )^2 + l/\sigma_{\min}(A)^2\\
\geq & ~ l/\sigma_{\min}(A)^2
\end{align*}
where the first step follows from simple algebra, the second step follows from replacing $\sinh(Ax)$ with $z = \sinh(Ax)$ and $\cosh^2() = \sinh^2()+1$ (Fact~\ref{fac:e_cosh_sinh_exact}), the third step follows from $w_{i}^2 > 0.5b_{i}^2 + 1/\sigma_{\min}(A)^2-1$, the fourth step follows from simple algebra, the fifth step follows from $x^2\geq0, \forall x$.

Since we know $D_{i,i} > l$ for all $i \in [n]$ and Lemma~\ref{lem:ADA_pd}, we have 
\begin{align*}
A^\top D A \succeq (\min_{i \in [n]} D_{i,i}) \cdot \sigma_{\min}(A)^2 I_d \succeq l \cdot I_d
\end{align*}
Thus, Hessian is positive definite forever and thus the function is convex.
\end{proof}

\subsection{Hessian is Lipschitz}\label{sec:sinh:lipschitz}
\begin{lemma}[Hessian is Lipschitz]\label{lem:hessian_is_lipschitz_sinh}
If the following condition holds
 
\begin{itemize}
    \item Let $H(x) = \frac{\d^2 L_{\sinh,\reg}}{\d x^2}$
    \item Let $R > 2$
    \item $\|x \|_2 \leq R, \| y \|_2 \leq R$
    \item $\| A (x-y) \|_{\infty} < 0.01$
    \item $\| A \| \leq R$
    \item $\| b \|_2 \leq R$
\end{itemize}
Then we have
    \begin{align*}
        \| H(x) - H(y) \| \leq \sqrt{n} \exp(6R^2) \cdot \| x- y \|_2
    \end{align*}
\end{lemma}
\begin{proof}

We have
\begin{align}\label{eq:rewrite_H_diff_sinh}
& ~ \| H(x) - H(y) \| \notag \\
= & ~ \| A^\top \diag(2 \sinh(Ax) - b) \diag(\sinh(Ax)) A -  A^\top \diag(2 \sinh(Ay) - b) \diag(\sinh(Ay)) A \| \notag \\
\leq & ~ \| A \|^2 \cdot \|  (2 \sinh(Ax) - b) \circ \sinh(Ax) - (2 \sinh(Ay) - b) \circ \sinh(Ay) \|_2  \notag \\
= & ~ \| A \|^2 \cdot \| 2 (\sinh(Ax) + \sinh(Ay) )\circ ( \sinh(Ax) - \sinh(Ay) ) - b \circ ( \sinh(Ax) - \sinh(A y) ) \|_2 \notag \\
= & ~ \| A \|^2 \cdot \| ( 2 \sinh(Ax) + 2 \sinh(Ay) - b ) \circ ( \sinh(Ax) - \sinh(Ay) ) \|_2  \notag \\
\leq & ~ \| A \|^2 \cdot \| ( 2 \sinh(Ax) + 2 \sinh(Ay) - b ) \|_{\infty} \cdot \| \sinh(Ax) - \sinh(Ay) \|_2
\end{align}
where the first step follows from $H(x) = \nabla^2L$ and simple algebra, 
the second step follows from Fact~\ref{fac:matrix_norm}, the third step follows from simple algebra, the fourth step follows from simple algebra, the last step follows from Fact~\ref{fac:vector_norm}.

For the first term in Eq.~\eqref{eq:rewrite_H_diff_sinh}, we have
\begin{align}\label{eq:upper_bound_H_x_H_y_step_1_sinh}
\| A \|^2 \leq R^2
\end{align}

For the second term in Eq.~\eqref{eq:rewrite_H_diff_sinh}, we have
\begin{align}\label{eq:upper_bound_H_x_H_y_step_2_sinh}
\| ( 2 \sinh(Ax) + 2 \sinh(Ay) - b ) \|_{\infty} \notag
\leq & ~ \| 2 \sinh(Ax)\|_{\infty} + \|2 \sinh(Ay)\|_{\infty} + \|b\|_\infty \notag \\
\leq & ~ \| 2 \cosh(Ax)\|_{\infty} + \|2 \cosh(Ay)\|_{\infty} + \|b\|_\infty \notag \\
\leq & ~ 2\exp(\|Ax\|_2) + 2\exp(\|Ay\|_2) + \|b\|_\infty \notag \\
\leq & ~ 4 \exp(R^2) + \| b \|_{\infty} \notag \\
\leq & ~ 4 \exp(R^2) + R \notag \\
\leq & ~ 5 \exp(R^2)
\end{align}

where the first step follows from Fact~\ref{fac:vector_norm}
, the second step follows from Fact~\ref{fac:vector_norm}, the third step follows from Fact~\ref{fac:vector_norm},  the fourth step follows from $\|Ax\|_2 \leq R^2, \|Ay\|_2 \leq R^2$, the fifth step follows from $\|b\|_\infty \leq R$, the last step follows from $R \geq 2$.

For the third term in Eq.~\eqref{eq:rewrite_H_diff_sinh}, we have
\begin{align}\label{eq:upper_bound_H_x_H_y_step_3_sinh}
\| \sinh(Ax) - \sinh(Ay) \|_2 
\leq & ~ \| \cosh(Ax) \|_2 \cdot 2 \| A (y-x) \|_{\infty} \notag \\
\leq & ~ \sqrt{n}\| \cosh(Ax) \|_\infty \cdot 2 \| A (y-x) \|_{\infty} \notag \\
\leq & ~ \sqrt{n}\exp(\|Ax\|_2) \cdot 2 \| A (y-x) \|_{2} \notag \\
\leq & ~ \sqrt{n}\exp(R^2)  \cdot 2 \| A (y-x) \|_2 \notag \\
\leq & ~ \sqrt{n}\exp(R^2)  \cdot 2 \| A \| \cdot \| y - x \|_2 \notag\\
\leq & ~ 2 \sqrt{n}R \exp(R^2) \cdot \|y - x\|_2
\end{align}
where the first step follows  from $\| A (y-x) \|_{\infty} < 0.01$ and Fact~\ref{fac:vector_norm}, the second step follows from Fact~\ref{fac:vector_norm}, the third step follows from Fact~\ref{fac:vector_norm},   the fourth step follows from $\|Ax\|_2 \leq R^2$, the fifth step follows from Fact~\ref{fac:matrix_norm}, the last step follows from $\|A\| \leq R$.

Putting it all together, we have
\begin{align*}
 \| H(x) - H(y) \| 
 \leq & ~ R^2 \cdot 5 \exp(R^2) \cdot 2 \sqrt{n} R\exp(R^2) \| y - x \|_2 \\
= & ~ 10 \sqrt{n} R^3 \exp(2R^2) \cdot \| y - x \|_2 \\
 \leq & ~ \sqrt{n}\exp(4R^2) \cdot  \exp(2R^2) \cdot \| y - x \|_2 \\
 = & ~ \sqrt{n}\exp(6R^2) \cdot \| y - x \|_2
\end{align*}
where the first step follows from by applying Eq.~\eqref{eq:upper_bound_H_x_H_y_step_1_sinh}, Eq.~\eqref{eq:upper_bound_H_x_H_y_step_2_sinh}, and Eq.~\eqref{eq:upper_bound_H_x_H_y_step_3_sinh}, the second step follows from simple algebra, the third step follows from $R \geq 2$, the last step follows from simple algebra.

\end{proof}

%% file: newton.tex
\section{Newton Method}\label{sec:newton}
 
In this section, we provide an approximate version of the Newton method for solving convex optimization problems and provide a detailed analysis of such a method. In Section~\ref{sec:newton:definitions} we define some assumptions under which we can tackle the optimization problem efficiently. In Section~\ref{sec:newton:connection} we state a simple lemma which is useful in Section~\ref{sec:newton:shrink}.
In Section~\ref{sec:newton:approximation} we provide an approximation variant for the update step of the newton method for convex optimization. In Section~\ref{sec:newton:hessian_property} we provide the upper bound of $\|H(x_k)\|$.  In Section~\ref{sec:newton:shrink} we provide the upper bound for $\|r_{k+1}\|$ and thus showed that our approximate update step is effective in solving the optimization problem. In Section~\ref{sec:newton:induction} we provide a lemma that showed our update step is effective. In Section~\ref{sec:newton:main}, we prove our main result.

\subsection{Definition and Update Rule}\label{sec:newton:definitions}
Let us study the local convergence of the Newton method. Consider the problem
\begin{align*}
    \min_{x \in \R^d } f(x)
\end{align*}
under the following assumptions:
\begin{definition}\label{def:f_ass}
We have
\begin{itemize}
    \item {\bf $l$-local Minimum.} Let $l > 0$ denote a parameter. There is a vector $x^* \in \R^d$ such that
    \begin{itemize}
        \item $\nabla f(x^*) = {\bf 0}_d$.
        \item $\nabla^2 f(x^*) \succeq l \cdot I_d$.
    \end{itemize}
    \item {\bf Hessian is $M$-Lipschitz.} Let $M>0$ denote a parameter that  \begin{align*}
        \| \nabla^2 f(y) - \nabla^2 f(x) \| \leq M \cdot \| y - x \|_2 
    \end{align*}
    \item {\bf Good Initialization Point.} Let $r_0:=\| x_0 -x_*\|_2$ such that
    \begin{align*}
        r_0 M \leq 0.1 l
    \end{align*}    
\end{itemize}
\end{definition}

We define gradient and Hessian as follows
\begin{definition}[Gradient and Hessian]=
We define gradient function $g : \R^d \rightarrow \R^d$ as
\begin{align*}
    g(x) := \nabla f(x)
\end{align*}
We define the Hessian function $H : \R^d \rightarrow \R^{d \times d}$ ,
\begin{align*}
    H(x) := \nabla^2 f(x)
\end{align*}
\end{definition}

Using the $g: \R^d \rightarrow \R^d$ and $H : \R^d \rightarrow \R^{d \times d}$, we can rewrite the exact process as follows 
:
\begin{definition}[Exact update]\label{def:exact_update_variant}
\begin{align*}
    x_{k+1} = x_k - H(x_k)^{-1} \cdot g(x_k)
\end{align*}
\end{definition}
\subsection{Connection between Gradient and Hessian}\label{sec:newton:connection}

\begin{lemma}[folklore]\label{lem:integral_gradient_hessian}
Let $g$ denote the gradient function and let $H: \R^d \rightarrow \R^{d \times d}$ denote the hessian function, then for any $x,y$, we have
\begin{align*}
g(y) - g(x) = \int_0^1 H( x + \tau(y-x) ) \cdot (y-x) \d \tau
\end{align*}
\end{lemma}
\begin{proof}

We have
\begin{align*}
\int_0^1 H( x + \tau(y-x) ) \cdot (y-x) \d \tau
= & ~ g( x+ \tau(y-x) ) \big|_0^1 \\
= & ~ g(x + 1\cdot (y-x)) - g(x+ 0 \cdot(y-x) ) \\
= & ~ g(y) - g(x)
\end{align*}
\end{proof}

\subsection{Approximation of Hessian and Update Rule}\label{sec:newton:approximation}
In many optimization applications, computing $\nabla^2 f(x_k )$ or $(\nabla^2 f(x_k))^{-1}$ is quite expensive. Therefore, a natural motivation is to approximately formulate its Hessian or inverse of Hessian.

\begin{definition}[Approximate Hessian]\label{def:wt_H}
For any $H(x_k)$, we define $\wt{H}(x_k)$ to satisfy the following condition 
\begin{align*}
 (1-\epsilon_H) \cdot H(x_k) \preceq \wt{H}(x_k) \preceq (1+\epsilon_H) \cdot H(x_k) .
\end{align*}
\end{definition}

To efficiently compute $\wt{H}(x_k)$, we use a standard tool from the literature
\begin{lemma}[\cite{syyz22,dsw22}]\label{lem:subsample}
Let $\epsilon_H = 0.01$. 
Given a matrix $A \in \R^{n \times d}$, for any positive diagonal matrix $D \in \R^{n \times n}$, there is an algorithm that runs in time
\begin{align*}
O( (\nnz(A) + d^{\omega} ) \poly(\log(n/\delta)) )
\end{align*}
output a $O(d \log(n/\delta))$ sparse diagonal matrix $\wt{D} \in \R^{n \times n}$ such that 
\begin{align*}
(1- \epsilon_H) A^\top D A \preceq A^\top \wt{D} A \preceq (1+\epsilon_H) A^\top D A.
\end{align*}
\end{lemma}

\begin{definition}[Approximate update]\label{def:update_x_k+1}
We consider the following process
\begin{align*}
    \underbrace{ x_{k+1} }_{d \times 1} = \underbrace{ x_k }_{d \times 1} - \underbrace{ \wt{H}(x_k)^{-1} }_{d \times d} \cdot \underbrace{ g(x_k) }_{d \times 1}
\end{align*}
\end{definition}

\subsection{Property of Hessian}\label{sec:newton:hessian_property}
\begin{lemma}\label{lem:lower_bound_spectral_hessian}
If the following conditions hold
\begin{itemize}
    \item Let $f$ be function that Hessian is $M$-Lipschitz (see Definition~\ref{def:f_ass}) 
    \item Suppose the optimal solution $x^*$  satisfy that $\| H(x_*) \| \geq l$ (see Definition~\ref{def:f_ass})
    \item Let $r_k := \| x_k - x^* \|_2$
\end{itemize}
 
We have
\begin{align*}
\| H(x_k) \| \geq l - M \cdot r_k
\end{align*}
\end{lemma}
\begin{proof}
We can show that
\begin{align*}
\| H(x_k) \| \geq & ~ \| H(x_*) \| - \| H(x_*) - H(x_k) \| \\
\geq & ~ \| H(x_*) \| - M \cdot \| x_* - x_k \|_2 \\
\geq & ~ l - M r_k
\end{align*}
where the first step follows from Fact~\ref{fac:matrix_norm}, the second step follows from $f$ is a $M$-bounded function, the third step follows from $\|H(x_k)\|\geq l$.
\end{proof}

\subsection{One Step Shrinking Lemma}\label{sec:newton:shrink}
\begin{lemma}\label{lem:one_step_shrinking}
If the following condition hold
\begin{itemize}
    \item Function $f$ follows from Definition~\ref{def:f_ass}. 
    \item Let $H$ denote the Hessian of $f$
    \item Let $g$ denote the gradient of $f$
    \item Let $r_k:= \| x_k - x^* \|_2$
\end{itemize}
 Then we have
\begin{align*}
r_{k+1} \leq 2(\epsilon_H + \frac{M r_k}{ l - M r_k } ) \cdot r_k
\end{align*} 
\end{lemma}

\begin{proof}
We have
\begin{align}\label{eq:variant_of_r_k_plus_1}
    x_{k+1} - x^* 
    = & ~ x_k - x^* - \wt{H}(x_k)^{-1} \cdot g(x_k)  \notag \\
    = & ~ x_k - x^* - \wt{H}(x_k)^{-1} \cdot ( g(x_k) - g(x^*) ) \notag\\ 
    = & ~ x_k - x^* - \wt{H}(x_k)^{-1} \cdot \int_0^1 H ( x^* + \tau (x_k-x^*) ) (x_k - x^*) \d \tau \notag\\ 
    = & ~ \wt{H}(x_k)^{-1} ( \wt{H}(x_k) ( x_k - x^*) ) - \wt{H}(x_k)^{-1} \cdot \int_0^1 H ( x^* + \tau (x_k-x^*) ) (x_k - x^*) \d \tau \notag\\ 
    = & ~  \wt{H}(x_k)^{-1} \bigg(\wt{H}(x_k) - \int_0^1 H ( x^* + \tau (x_k-x^*) ) \d \tau \bigg) \cdot (x_k - x^*) \notag\\ 
    = & ~ \wt{H}(x_k)^{-1} \bigg(\int_0^1 \wt{H}(x_k) \d \tau - \int_0^1 H ( x^* + \tau (x_k-x^*) ) \d \tau \bigg) \cdot (x_k - x^*) \notag\\ 
    = & ~  \bigg(\wt{H}(x_k)^{-1} \int_0^1 (\wt{H}(x_k) - H ( x^* + \tau (x_k-x^*)) \d \tau \bigg) \cdot (x_k - x^*) \notag\\ 
    = & ~ G_k \cdot (x_k - x^*)
\end{align}
where the first step follows from Definition~\ref{def:update_x_k+1}, the second step follows from $g(x^*) = {\bf 0}_d$, the third step follows from  Lemma~\ref{lem:integral_gradient_hessian}, the forth step follows from $H^{-1} H = I$, the fifth step follows from simple algebra, the sixth step follows from simple algebra, 
 the last step follows from rewrite the equation using $G_k$ below
:
\begin{align*}
    G_k := \wt{H}(x_k)^{-1} \cdot \int_0^1 \wt{H}(x_k) - H(x^*+ \tau (x_k - x^*)) \d \tau
\end{align*}

Then we can bound $\| G_k \|$ as follows
\begin{align}\label{eq:upper_bound_G_k_divide}
    \| G_k \|
    = & ~ \Big\| \wt{H}(x_k)^{-1} \cdot  \int_0^1 ( \wt{H}(x_k) - H( x^* + \tau (x_k - x^*) ) ) \d \tau \Big\| \notag \\ 
    = & ~ \Big\| \wt{H}(x_k)^{-1} \cdot  \int_0^1 ( \wt{H}(x_k) - H(x_k) + H(x_k) - H( x^* + \tau (x_k - x^*) ) ) \d \tau \Big\| \notag \\
    = & ~ \Big\| \wt{H}(x_k)^{-1} \cdot  \Big(\int_0^1 ( \wt{H}(x_k) - H(x_k)) \d \tau + \int_0^1(H(x_k) - H( x^* + \tau (x_k - x^*) ) ) \d \tau \Big) \Big\| \notag \\
    \leq & ~\Big\| \wt{H}(x_k)^{-1} \cdot \int_0^1 ( \wt{H}(x_k) - H(x_k) ) \d \tau \Big\| + \Big\| \wt{H}(x_k)^{-1} \cdot \int_0^1 (H(x_k) - H( x^* + \tau (x_k - x^*) )) \d \tau \Big\|
\end{align}
where the first step follows from the definition of $G_k$, the second step follows from simple algebra, the third step follows from simple algebra, the last step follows from Fact~\ref{fac:matrix_norm}.

For the first term, we have

\begin{align}\label{eq:upper_bound_G_k_1st_step}
    \Big\| \wt{H}(x_k)^{-1} \cdot \int_0^1 ( \wt{H}(x_k) - H(x_k) ) \d \tau \Big\| \notag
    = & ~ \Big\| \wt{H}(x_k)^{-1} \cdot ( \wt{H}(x_k) - H(x_k) ) \int_0^1 \d \tau \Big\| \\ \notag
    = & ~ \| \wt{H}(x_k)^{-1}  ( \wt{H}(x_k) - H(x_k) ) \|  \\
    \leq & ~ 2\epsilon_H
\end{align}
where the first step follows from simple algebra, the second step follows from $\int_0^1 \d \tau = 1$, the third step follows from Fact~\ref{fac:psd}.

For the second term, we have
\begin{align}\label{eq:upper_bound_G_k_2nd_step}
    & ~\Big\| \wt{H}(x_k)^{-1} \cdot \int_0^1 (H(x_k) - H( x^* + \tau (x_k - x^*) ))  \d \tau \Big\|  \notag \\
    \leq & ~ \| \wt{H}(x_k)^{-1} \| \cdot \Big\| \int_0^1 (H(x_k) - H( x^* + \tau (x_k - x^*) ) ) \d \tau \Big\| \notag \\ 
    \leq & ~ (1+\epsilon_H) \cdot \| H(x_k)^{-1} \| \cdot \Big\| \int_0^1 (H(x_k) - H( x^* + \tau (x_k - x^*) ) )  \d \tau \Big\| \notag\\ 
    \leq & ~ (1+\epsilon_H) \cdot \| H(x_k)^{-1} \| \cdot \int_0^1 \Big\|  H(x_k) - H( x^* + \tau (x_k - x^*) )  \Big\| \d \tau \notag\\ 
    \leq & ~ (1+\epsilon_H) \cdot \| H(x_k)^{-1} \| \cdot \max_{\tau \in [0,1]} \Big\|  H(x_k) - H( x^* + \tau (x_k - x^*) )  \Big\| \notag\\ 
    \leq & ~ (1+\epsilon_H) \cdot \| H(x_k)^{-1} \| \cdot r_k M \notag\\ 
    \leq & ~ (1+\epsilon_H) \cdot (l - M r_k)^{-1} \cdot r_k M \notag\\
    \leq & ~ 2 \frac{M r_k}{l- M r_k}
\end{align}
where the first step follows from Fact~\ref{fac:matrix_norm}, the second step follows from $(1+\epsilon_H)H(x_k)<\wt{H}(x_k)$ and Fact~\ref{fac:matrix_norm}
, the third step follows from $\|\int \d \tau \| \leq \int \| \| \d \tau$, the forth step follows from $\int_0^1 f(\tau) \d \tau \leq \max_{\tau \in [0,1]} f(\tau)$
, the fifth step follows from Definition~\ref{def:f_ass}
, the sixth step follows from $\|H(x_k)\| \ge l-Mr_k$ (see Lemma~\ref{lem:lower_bound_spectral_hessian}), the last step follows from $\epsilon_H \in (0,1)$.

Thus, we have, 
    \begin{align*}
        r_{k+1}
        = & ~ \|G_k \cdot (x_k - x^*)\| \\
        \leq & ~ \|G_k\| \cdot \|(x_k - x^*)\| \\
        = & ~ \|G_k\| \cdot r_k \\
        \leq & ~ 2(\epsilon_H +  \frac{M r_k}{l- M r_k}) \cdot r_k
    \end{align*}

where the first step follows from Eq.~\eqref{eq:variant_of_r_k_plus_1} and by definition of $r_k$, the second step follows form $\|ab\| \leq \|a\|\|b\|, \forall a,b$, the third step follows from $r_k = \|x_k - x^*\|$, the last step follows from Eq.~\eqref{eq:upper_bound_G_k_divide}, Eq.~\eqref{eq:upper_bound_G_k_1st_step}, and Eq.~\eqref{eq:upper_bound_G_k_2nd_step}.

\end{proof}

\subsection{Induction}\label{sec:newton:induction}

\begin{lemma}\label{lem:newton_induction}
If the following condition hold
\begin{itemize}
    \item $\epsilon_H = 0.01$
    \item $r_{i} \leq 0.4 r_{i-1}$, for all $i \in [k]$
    \item $M \cdot r_i \leq 0.1 l$, for all $i \in [k]$
\end{itemize}
Then we have
\begin{itemize}
    \item $r_{k+1} \leq 0.4 r_k$
    \item $M \cdot r_{k+1} \leq 0.1 l$
\end{itemize}
\end{lemma}
\begin{proof}

{\bf Proof of Part 1.}

 We have
 \begin{align*}
r_{k+1} \leq &  2(\epsilon_H + \frac{M r_k}{ l - M r_k } ) \cdot r_k \\
\leq & ~ 2 ( 0.01 + \frac{M r_k}{ l - M r_k }  ) r_k \\
\leq & ~ 2 (0.01 + \frac{0.1 l}{l - 0.1l}) r_k \\
\leq & ~ 0.4 r_k
 \end{align*}
 where the first step follows from Lemma~\ref{lem:one_step_shrinking}.

{\bf Proof of Part 2.}

We have 
\begin{align*}
M \cdot r_{k+1} 
\leq & ~ M \cdot 0.4 r_{k} \\
\leq & ~ 0.4 \cdot 0.1 l \\
\leq & ~ 0.1 l
\end{align*}

\end{proof}

\subsection{Main Result}\label{sec:newton:main}

We state our main result as follows.
\begin{theorem}[Formal version of Theorem~\ref{thm:main_informal}]\label{thm:main_formal}
Given matrix $A \in \R^{n \times d}$, $b \in \R^n$, and $w \in \R^n$. 

Let $f$ be any of functions $\exp, \cosh$ and $\sinh$.

Let $x^*$ denote the optimal solution of 
\begin{align*}
\min_{x \in \R^d} 0.5 \| f(Ax) - b \|_2^2 + 0.5 \| \diag(w) A x \|_2^2
\end{align*}
that $g(x^*) = {\bf 0}_d$ and $\| x^* \|_2 \leq R$.

Let $\| A \| \leq R, \| b \|_2 \leq R$.

\begin{itemize}
    \item Let $w_{i}^2 \geq 0.5 b_i^2 + l$ for all $i \in [n]$. (If $f=\exp$, see Lemma~\ref{lem:hessian_is_pd_exp})
    \item Let $w_{i}^2 \geq 0.5 b_i^2 + l/\sigma_{\min}(A)^2 + 1$ for all $i \in [n]$. (If $f=\cosh$, see Lemma~\ref{lem:hessian_is_pd_cosh})
    \item Let $w_{i}^2 \geq 0.5 b_i^2 + l/\sigma_{\min}(A)^2 - 1$ for all $i \in [n]$. (If $f=\sinh$, see Lemma~\ref{lem:hessian_is_pd_sinh})
    
\end{itemize}

Let $M = \sqrt{n} \cdot \exp(6R^2)$.

Let $x_0$ denote an initial point such that $M \| x_0 - x^* \|_2 \leq 0.1 l$.

For any accuracy parameter $\epsilon \in (0,0.1)$ and failure probability $\delta \in (0,0.1)$.  There is a randomized algorithm (Algorithm~\ref{alg:main}) that runs $\log(\| x_0 - x^* \|_2/ \epsilon)$ iterations and spend  
\begin{align*}
O( (\nnz(A) + d^{\omega} ) \cdot \poly(\log(n/\delta)) 
\end{align*}
time per iteration, and finally outputs a vector $\wt{x} \in \R^d$ such that
\begin{align*}
\| \wt{x} - x^* \|_2 \leq \epsilon
\end{align*}
holds with probability at least $1-\delta$.
 
\end{theorem}
\begin{proof}
{\bf Proof of $\exp$ function.}

It follows from combining Lemma~\ref{lem:hessian_is_pd_exp}, Lemma~\ref{lem:newton_induction}, Lemma~\ref{lem:subsample}, and
Lemma~\ref{lem:one_step_shrinking}.

After $T$ iterations, we have
\begin{align*}
\| x_T - x^* \|_2 \leq 0.4^T \cdot \| x_0 - x^* \|_2
\end{align*}
By choice of $T$, we get the desired bound. The failure probability is following from union bound over $T$ iterations.

{\bf Proof of $\cosh$ function.}

It follows from combining Lemma~\ref{lem:hessian_is_pd_cosh}, Lemma~\ref{lem:newton_induction}, Lemma~\ref{lem:subsample}, and
Lemma~\ref{lem:one_step_shrinking}.

{\bf Proof of $\sinh$ function.}

It follows from combining Lemma~\ref{lem:hessian_is_pd_sinh}, Lemma~\ref{lem:newton_induction}, Lemma~\ref{lem:subsample}, and
Lemma~\ref{lem:one_step_shrinking}.

\end{proof}